\documentclass[10pt,twocolumn,letterpaper]{article}

\usepackage{iccv}              % To produce the CAMERA-READY version
\definecolor{iccvblue}{rgb}{0.21,0.49,0.74}
\usepackage[pagebackref,breaklinks,colorlinks,allcolors=iccvblue]{hyperref}

\def\paperID{13023} % *** Enter the Paper ID here
\def\confName{ICCV}
\def\confYear{2025}

\title{Segmentation Assisted Incremental Test Time Adaptation in an Open World}

\author{Manogna Sreenivas and Soma Biswas\\
Indian Institute of Science\\
Bengaluru, India\\
{\tt\small \{manognas,somabiswas\}@iisc.ac.in}
}

\usepackage{subcaption}
\usepackage{adjustbox}
\usepackage{pifont}% http://ctan.org/pkg/pifont
\usepackage{amsfonts}       % blackboard math symbols
\usepackage{multicol}
\usepackage{multirow}
\usepackage{pgfplots}
\usepackage{tikz}
\usepackage{xcolor}
\usepackage{threeparttable}
\usetikzlibrary{shapes.geometric, arrows, positioning}

\definecolor{seen}{HTML} {50AE8B}
\definecolor{unseen}{HTML}{F58C6C}

\newcommand{\cmark}{\ding{51}}%
\newcommand{\xmark}{\ding{55}}%

\begin{document}
\maketitle

\begin{abstract}
%    In this paper, we study the problem of open world test time adaptation (TTA) using Vision-Language Models (VLMs). Our method addresses the challenge of handling samples that arrive sequentially, with predefined classes, while also accommodating unknown samples that emerge during testing. We aim to incrementally incorporate these unknown samples into our set of desired classes in an online manner. To achieve this, we leverage active learning to query uncertain samples. If a queried sample belongs to a new class, it is added to the set of desired classes, and TTA continues. We establish baselines using several well-known active learning strategies and baseline TTA methods with CLIP. This work is the first to set up this practical TTA protocol. Additionally, we propose a novel end-to-end framework that integrates active learning and TTA, utilizing pixel-wise segmentation to further enhance the process.
In dynamic environments, unfamiliar objects and distribution shifts are often encountered which challenge the generalization abilities of the deployed trained models. 
This work addresses Incremental Test Time Adaptation (ITTA) of Vision-Language Models (VLMs), tackling scenarios where unseen classes and unseen domains continuously appear during testing. 
Unlike traditional Test Time Adaptation approaches, where the test stream comes only from a predefined set of classes, our framework allows models to adapt simultaneously to both covariate and label shifts, actively incorporating new classes as they emerge.
Towards this goal, we establish a new benchmark for ITTA, integrating single-image TTA methods for VLMs with active labeling techniques that query an oracle for samples potentially representing unseen classes during test time.
We propose a segmentation assisted active labeling module, termed SegAssist, which is training-free and repurposes the VLM’s segmentation capabilities to refine active sample selection, prioritizing samples likely to belong to unseen classes. Extensive experiments on several benchmark datasets demonstrate the potential of SegAssist to enhance the performance of VLMs in real-world scenarios, where continuous adaptation to emerging data is essential.
% as used by the base TTA model to query uncertain samples, thereby incorporating new classes dynamically as they are discovered.
% Leveraging our novel segmentation-assisted active learning module, existing TTA approaches can be equipped with the capability to effectively operate in an open-world.
% The proposed module, termed SegAssist is training-free and re-purposes the same VLM as used by the base TTA model to query uncertain samples, thereby incorporating new classes dynamically as they are discovered.  
% Extensive experiments on several benchmark datasets demonstrate the potential of SegAssist to enhance the performance of VLMs in real-world scenarios, where continous online adaptation to emerging data is essential.
\end{abstract}

\vspace{-15pt}
\section{Introduction}
\label{sec:intro}

% Consider a real scenario where a trained model is deployed for detecting cars, buses, etc primarily seen in urban environments suddenly encounters tractors, bullock carts, animals, etc. as it passes through a rural region.
% In this situation, the model must recognize these unexpected objects as new classes rather than misclassifying them as existing categories, and  should further integrate this new knowledge seamlessly in the model.
% This is in addition to correctly recognizing the known objects in  considerably different lighting, weather conditions.
Consider a model deployed in real world, navigating a dynamic environment where it encounters familiar objects in unfamiliar settings and new objects it has never seen before. For instance, a model trained to recognize urban entities like cars, trucks, and pedestrians might suddenly come across unfamiliar elements, such as bicycles or animals, when it transitions to a rural landscape. In these situations, the model must not only adapt to the changed appearance of known objects, affected by varying factors like lighting and weather, but also correctly identify unfamiliar objects as new classes instead of misclassifying them. Moreover, it must seamlessly integrate this new information to remain accurate and relevant in real-time.
This is the underlying motivation of {\em Incremental Test Time Adaptation (ITTA)}, where a deployed model must continuously adapt to changes in data distribution and incorporate emerging classes in an online manner. 
Traditional TTA methods~\cite{tent, santa, pstarc, dss} aim to recognize known classes in unseen domains with batches of test data. Recently, VLMs such as CLIP~\cite{clip}, have shown excellent zero-shot generalization abilities due to their pre-training on web-scale image-text data, making them promising candidates for the challenging single-image TTA scenario~\cite{tpt, tda}.

In this work, we propose a more realistic Incremental TTA setting, since data in the open-world can come not only from unseen distributions, but from unseen classes as well.  
% During testing, as each test sample is encountered, the model identifies and categorizes familiar objects, while actively seeking queries for uncertain samples to discover and incorporate new classes, thereby continuously expanding its knowledge in an online setting.
Our protocol reimagines TTA as an evolving learning process, suited for real-world applications where environments are constantly changing. This is done by actively querying an oracle for labels of uncertain samples to discover and incorporate new classes, thereby continuously expanding its knowledge in an online setting.
Towards this goal, we propose a novel active sample selection module termed SegAssist, which leverages pixel-wise segmentation to determine whether an uncertain sample should be chosen for active labeling.
This ensures that we utilize the limited active labeling budget judiciously, primarily for labeling samples of new classes.
{\em SegAssist is a training-free plug-in module which can be seamlessly integrated with existing single-sample TTA frameworks} to adapt them to the ITTA scenario. 
Our key contributions are summarized as follows:
\begin{itemize}
    \item We address the real-world challenging task of Incremental Test Time Adaptation (ITTA) in an open-world with both covariate and label shift using VLMs.
    \item We establish strong baselines using different active labeling strategies and state-of-the-art CLIP-based TTA methods, paving the way for future research in this direction. 
    \item We propose a novel training-free active labeling module, named SegAssist, which dynamically queries and incorporates new classes, supporting class recognition in a continuously expanding set of classes in an online setting. SegAssist can be seamlessly plugged in with existing TTA methods to equip them for ITTA. 
    \item We propose a novel metric termed Incremental Class Detection Delay (ICDD) to quantify an ITTA method in terms of the timeliness of detecting new classes. 
    \item We demonstrate the effectiveness of our method through extensive experiments on several benchmark datasets, showcasing its robustness in handling both covariate and label shifts during test time. 
\end{itemize}
\section{ITTA: A step towards TTA in real world}
\label{sec:motivation}

\begin{table}[]
    \begin{adjustbox}{max width=\linewidth}
    \begin{tabular}{@{}ccccccc@{}}
    \toprule
    \multirow{2}{*}{Training} & \multirow{2}{*}{TTA} & Covariate &  Label  & Single  & Classification  & \multirow{2}{*}{Methods}   \\
                              &                      &   shift   &  shift  &  image  &      task       &                            \\ \midrule
            \cmark            &         \cmark       &   \cmark  &  \xmark & \xmark  &     $C_s$       &   TENT\cite{tent}, CoTTA\cite{cotta}, ROID\cite{roid}         \\
            \xmark            &         \cmark       &   \cmark  &  \xmark & \cmark  &     $C_s$       &   TPT\cite{tpt}, TDA\cite{tda}, DPE\cite{dpe}         \\
            \xmark            &         \cmark       &   \cmark  &  \cmark & \xmark  &   $C_s + 1$   &     Open world TTA~\cite{wofcrowds, dproto}             \\
            \cmark            &         \xmark       &   \xmark  &  \cmark & \xmark  &   $C_s$+$C_u$   &     Class Incremental Learning~\cite{lwf, icarl, ucir}             \\
            \xmark            &         \cmark       &   \cmark  &  \cmark & \cmark  &   $C_s$+$C_u$   &            Proposed ITTA framework             \\ \bottomrule

    \end{tabular}
    \end{adjustbox}
\caption{Comparison of the proposed ITTA framework with existing research directions.}
\label{diff-methods}
\vspace{-10pt}
\end{table}

The motivation for addressing Incremental Test Time Adaptation (ITTA) stems from the growing demand for models that can handle dynamically changing, real-world environments, where assumptions of a fixed distribution between training and test data rarely hold. Existing approaches have attempted to address isolated aspects of these challenges, yet they often fall short in real-world scenarios such as autonomous systems, surveillance, etc. which demand that models adapt to changes in both the nature of input data (covariate shift) and the introduction and assimilation of entirely new object categories (label shift).
We first describe the existing closely related research directions (Table~\ref{diff-methods}), motivating the need for the proposed ITTA framework.  
%that combine multiple, complex conditions like covariate shift, label shift and incremental learning.
%Real-world applications such as autonomous systems, surveillance, and evolving content moderation demand that models adapt to changes in both the nature of input data (covariate shift) and the introduction of entirely new categories of objects, behaviors, or labels (label shift).
%In real-world scenarios, models must not only deal with covariate shifts (i.e., differences in data distribution, such as lighting or weather conditions) but also adapt to label shifts where previously unseen classes are introduced.

\begin{itemize}
\item {\bf TTA with only Covariate Shift:} Traditional TTA methods like TENT~\cite{tent}, CoTTA~\cite{cotta}, ROID~\cite{roid} focus on adapting models trained on clean data to handle changes in data distribution during test-time without changing the underlying classes. 
{\em These methods cannot accommodate dynamically emerging new classes.}

\item {\bf Single-Image TTA with Covariate Shift:} With the advent of pretrained VLMs, methods like TPT~\cite{tpt}, TDA~\cite{tda}, DPE~\cite{dpe} address single-image TTA, which is closer to a real-time scenario, but with   fixed label space. 
{\em Thus, they also cannot handle test samples from unseen classes outside the $C_s$ predefined classes of interest.}
\item {\bf Open-World TTA with Covariate and Label Shift:} To handle unknown classes, only recently, open-world TTA methods~\cite{wofcrowds, dproto, rosita} propose to do $(C_s+1)$-way classification, where $C_s$ represents the seen classes and an additional class is reserved for all unknown classes. 
{\em However, these methods do not support incremental learning of these new classes, which is crucial in real-world.}
\item {\bf Class Incremental Learning (CIL):} Traditional CIL techniques~\cite{lwf, icarl} aim to incorporate new classes with time. However, they often assume access to training data for these new classes to continually learn the model. 
{\em Hence, they are ill-suited to dynamic TTA settings where only one unlabeled test sample is available at a time and immediate adaptation is desired.}
\end{itemize}
Our work combines these considerations by establishing a novel {\em Incremental Test Time Adaptation (ITTA)} protocol reflecting the online nature of real-world deployment.
ITTA requires the model to not only adapt to covariate shifts, but also incrementally learn new classes without relying on additional training data.
It enables the model to:
\begin{enumerate}
\item Continuously adapt to covariate shifts. %(covariate shift),
\item Recognize and incorporate new classes as they appear. %(label shift),
\item Operate on a single-image basis without access to batches or any labeled  data.
\item Maintain performance across seen classes and a growing set of  dynamically incorporated unseen classes.
\end{enumerate}
%The ITTA protocol addresses single-image adaptation, where test samples arrive one at a time, reflecting the online nature of real-world deployment. 
To address the need for continuous class expansion, we leverage active learning to query samples where the model exhibits uncertainty. When a queried sample belongs to a previously unseen class, we dynamically add this class to the list of seen classes, thereby enabling ongoing classification across both previously seen and emerging new categories. This framework enables efficient adaptation in highly variable, real-world environments. \\ \\
%, allowing the model to remain accurate and relevant over time.
% *******************
\noindent{\bf Vision Language Models for ITTA: }
VLMs like CLIP~\cite{clip} provide a flexible framework for incremental adaptation in open-world settings due to their ability to leverage predefined class names as text-based classifiers. In VLMs such as CLIP, classification is performed by matching the visual embedding of an image to the text embeddings of class names. 
Given a set of known classes (i.e. classes expected to be encountered for the given application) $\mathcal{C}_s = \{c_1, c_2, \dots, c_{S}\}$, where each class $c_i$ is associated with a descriptive label (e.g., “a photo of a car”), VLMs generate text embeddings $\phi_T(c_i)$ for each class $c_i$. These text embeddings act as the classifiers, enabling the model to perform zero-shot classification by computing the similarity between the image embedding $\phi_I(x_t)$ and each text embedding $\phi_T(c_i)$. 
Here $\phi_I$ and $\phi_T$ denote the image and text encoders of the VLM respectively.  

In the context of ITTA, VLMs are particularly advantageous due to their ability to dynamically expand the set of seen classes with minimal modification. During ITTA, incoming test samples are sequentially processed, and uncertain samples are flagged for active labeling queries. If a queried sample $x_t$ is identified as an unseen class $u$, it is added to the set of seen classes. Thanks to the text encoder, the classifier set can be updated to incorporate the text embedding of the newly discovered class.
%, updating $\mathcal{C}^t \leftarrow \mathcal{C}^{t+1} = \mathcal{C}^t \cup \{u\}$.
% The VLM-based framework then generates the text embedding $\phi_T(u)$ for this new class $u$ by embedding its descriptive label (e.g., “a photo of a tractor”). This text embedding is added to the existing set of text embeddings $\{\phi_T(c_i)\}_{i=1}^{C_{s}}$.
%Thus, the classifier in the VLM is incrementally expanded to accommodate new classes without requiring retraining on visual data, as the text embedding $\phi_T(u)$ for the new class can be obtained directly from the model’s pre-trained text encoder.
{\em This free classifier expansion mechanism is a key advantage of VLMs in ITTA, as it allows for seamless integration of new classes and enables the model to continuously adapt to new categories as they appear in an open-world setting.}

%\clearpage

%\subsection{Towards Real-World TTA}
%
%Difference between prior TTA scenarios and the proposed real-world TTA scenario. \ToDo{ \color{red} Visually illustrating would be good, but needs space.}
%
%\begin{figure}[h]
%           \centering
%           \begin{adjustbox}{max width = \linewidth}
%           \includegraphics{data/tta-scenarios}
%           \end{adjustbox}
%           \caption{Comparison of prior TTA scenarios and the proposed real-world TTA scenario.}
%           \label{fig:tta-scenarios}
%\end{figure}
%- We address the hardest scenario of all these.

%\subsection{Domain shifts in the context of CLIP}
%- Can't exactly say training and test domains are different as it is pretrained model. Performance on CIFAR-10C is worse compared to CIFAR-10. \ToDo{ \color{red} Add results. Not tried any experiments on corrupted datasets. It'd be good if it works for CIFAR-100. That would cover variety of domain shifts: ImageNet-R, DomainNet-Clipart, CIFAR-100C}
%
%\subsection{Incorporating new classes}
%- Once the new class is detected through active queryin, the text classifier can be extended by including the text embedding of the new class "A photo of a {new class name}". As we use VLM, we can get the new classifiers in an off-the-shelf manner. \ToDo{ \color{red} Mathematically write how text classifier is changed.}

\section{Related Works}
\label{sec:related_works}
Here, we describe some of the related works in literature. 

\noindent \textbf{Test Time Adaptation: }
TTA methods aim to adapt a model trained on one domain to test data from another domain without requiring access to the original training data. Early TTA approaches, such as TENT~\cite{tent}, BN Stats Adapt~\cite{bn_neurips}, CoTTA~\cite{cotta}, and ROID~\cite{roid} focus on adapting models to handle covariate shifts, such as changes in lighting, weather conditions and other common corruptions. These methods typically assume access to a batch of test samples at a time for continuous adaptation.
More recently, single-sample TTA based on pre-trained vision-language models like CLIP~\cite{clip} has been explored by approaches such as TPT~\cite{tpt}, TDA~\cite{tda}, and DPE~\cite{dpe}.
%For example, TPT utilizes prompt tuning and selective augmentation to minimize entropy, TDA employs dynamic adapters for pseudo-label refinement, and DPE leverages dual prototype evolving to accumulate task-specific knowledge from both text and visual modalities.
However, these methods address only covariate shifts and are not equipped to handle label shifts during testing, as in the realistic ITTA scenario which is of our interest.

%Recently, ~\cite{wofcrowds, dproto} have explored open-world TTA, where the model is updated such that it distinguish unknown class samples from known class samples. This is similar to the well studied OOD detection~\cite{clipn}, posed as a $(C_s+1)$-way classification problem, where $C_s$ represents the seen classes and a $+1$ class is reserved for the unknown class.

\noindent \textbf{Active Labeling (AL): }
AL methods aim to efficiently query informative samples to reduce the need for extensive labeling, making it particularly valuable in resource-constrained scenarios~\cite{active_learning}. Traditional AL methods~\cite{kmeans, activeda, clue} operate in an offline setting, typically requiring access to large, fixed datasets, and are applied during model training. Recently, AL has been explored in the TTA context, as in ATTA~\cite{atta}, which queries samples during test time. However,~\cite{atta} assumes that test samples arrive in batches and belong to a fixed set of classes, limiting their applicability in single image ITTA scenario.
In contrast to these works which query uncertain samples from the fixed set of classes, in this work, we propose to leverage active labeling to query uncertain samples during single-image test time adaptation to enable real-time incorporation of new classes as they appear. While many existing AL strategies rely on abundant data or batch processing~\cite{atta, activeda, clue, kmeans} to compute selection metrics, we explore selection criteria that can be calculated on a per-sample basis, suitable for single image TTA.

\noindent\textbf{Open World and Incremental Learning: }
Learning to recognize new classes in an open-world is a challenging problem, which has been studied from several perspectives. Out of Distribution (OOD) detection methods~\cite{msp, clipn, palm} aim to distinguish between seen and unseen class samples, typically formulated as a $(C_s+1)$-way classification problem, where $C_s$ represents the seen classes and a $+1$ class is reserved for the unknown classes. Traditional OOD approaches rely on access to abundant data and are generally applied during training~\cite{vos, npos, clipn, mos, palm}, although recent studies have explored OOD detection in a batch-wise TTA setting~\cite{wofcrowds, dproto}.
In contrast, incremental learning~\cite{lwf, icarl, ucir} address the challenge of adding new classes over time without forgetting previously learned classes. However, these methods assume access to training data for new classes and typically require retraining to incorporate new information, making them unsuitable for real-time TTA scenarios.

{\em These works address isolated aspects of real world scenarios, but they fall short in scenarios that combine multiple, complex conditions like covariate shift, label shift, and incremental learning.} Towards this goal, we introduce a practical protocol termed Incremental Test Time Adaptation (ITTA), designed to handle all these challenges simultaneously in an online, real-time manner. We establish strong baselines for this problem by combining single-sample active learning strategies with TTA methods using CLIP~\cite{clip}.

\section{Incremental Test Time Adaptation}

\begin{figure}[t]
           \centering
           \begin{adjustbox}{max width = \linewidth}
           \includegraphics{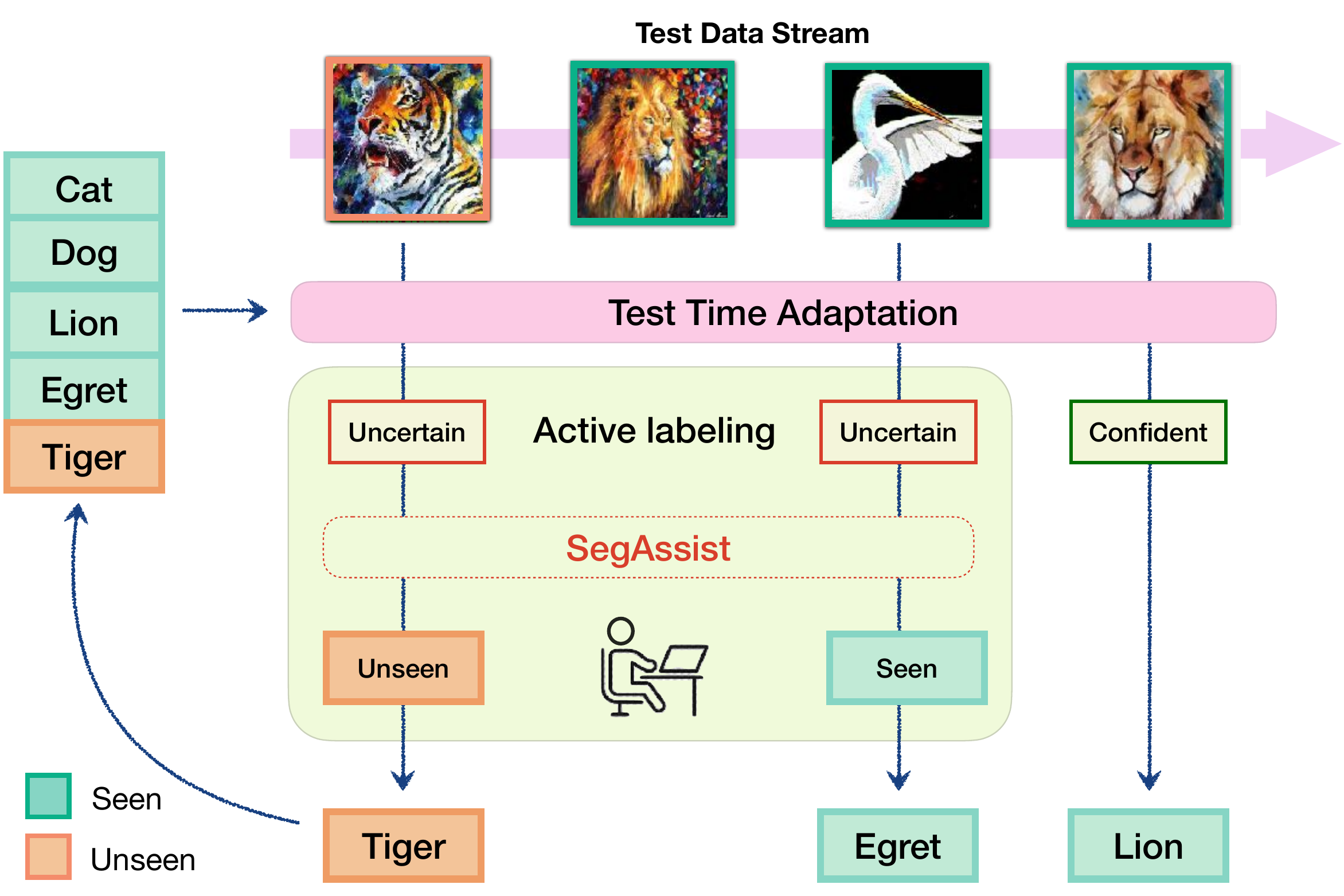}
           \end{adjustbox}
           \caption{ Incremental Test Time Adaptation: A test stream in real world can consist of samples with covariate (painting) and label shift (unseen classes). The goal here is to incorporate new classes by querying uncertain samples for active labeling. The class is added to the set of seen classes if the oracle identifies it as an unseen class. SegAssist is a plug-in module we propose to favour selection of unseen class samples for active labeling.}
           \label{fig:itta_framework}
           \vspace{-13pt}
\end{figure}

% Here, we formally describe the ITTA problem and then our proposed SegAssist active sample selection module.
\subsection{Problem Definition}
\label{sec:problem-definition}
In this work, we address Incremental Test-Time Adaptation (ITTA) of Vision-Language Models (VLMs) in an open-world, where both covariate and label shifts can occur.
During test-time adaptation, the model encounters a stream of test samples $\{x_t\}_{t=1, 2, \hdots}$, where each sample $x_t$ may either belong to the seen classes $C_s=\{c_1,c_2,\ldots c_S\}$ or a previously unseen class.
In the context of VLM, seen classes indicate the set of classes expected to be encountered in that scenario and for which labels are provided for classification. 
For example, in an autonomous driving scenario, the set of seen classes can be cars, buses, bikes, etc. 
The task is to predict a label $y_t$ for each incoming test sample $x_t$ while also adapting to distributional shifts and dynamically incorporating new classes as they emerge. This is done by integrating active labeling mechanism with TTA as shown in Fig.~\ref{fig:itta_framework}. 

We use the CLIP VLM consisting of an image encoder $\phi_{I}$ and text encoder $\phi_T$. Given the initial set of classes $C^{0}=C_s$, the text encoder $\phi_T$ is used to obtain the text classifiers by embedding text prompts of the form ``A photo of a $\{c_i\}$" for each class $c_i \in C^{0}$. For an incoming sample $x_t$ at time $t$, the model adapts itself and then predicts the label $y_t \in C^{t}$, where $C^t$ is the set of seen classes till time $t$.

% \ToDo{Define budget rate for active learning}

However, when the model encounters an uncertain sample, possibly a sample that does not belong to the set of seen classes $\mathcal{C}^t$, an active learning query is triggered. The goal is to label this uncertain sample and determine if it belongs to a new class. If the queried sample $x_t$ is confirmed to belong to a new, previously unseen class $u$, the model updates the set of known classes by adding the new class, i.e., $\mathcal{C}^{t+1} \leftarrow \mathcal{C}^t \cup \{u\}$ as illustrated in Fig.~\ref{fig:itta_framework}. Following this update, the model continues to classify subsequent samples from the expanded set of classes $\mathcal{C}^{t+1}$. 
%ensuring that the model can predict over a class set which grows incrementally as new classes are discovered during TTA.
The adaptation and prediction process continue at each time step. 
%with the model adjusting both to covariate shifts (new contexts for known classes) and label shifts (emerging new classes).
%The model’s class set $C_t$ grows dynamically, allowing it to incorporate both known and newly discovered classes as test samples are processed. This incremental adaptation ensures that the model remains robust and effective over time, even in open-world scenarios where the distribution of classes and contexts is constantly evolving.

\vspace{-10pt}
\paragraph{Budget for ITTA. }
\label{sec:budget-itta}
Prior active labeling works define the annotation budget as a percentage of the dataset size or a fixed number of samples that can be queried for labeling. Here, as the test samples are streamed in an online manner, we define a replenishing budget $\mathcal{B}_t$ more suited for an online scenario where we do not know the number of test samples apriori. $\mathcal{B}_t$ is set based on a budget rate $r$ (analogous to the percentage of data in offline active labeling) which is replenished after seeing every $N$ samples. 
% The budget available at time $t$ is updated as $\mathcal{B}(t; p,K) \leftarrow \mathcal{B}(t; p,K) + p.K$ at every $K$th time step, starting with an initial budget $\mathcal{B}(0; p, K) = p.K$. 
This budget controls the number of samples that can be queried for labeling. We set the budget rate $r$ to 1\% and $N$ to 1000, replenishing the budget by 10 queries every 1000 test samples seen.   

% Thus, the core objective of the ITTA framework is to dynamically and incrementally expand the set of classes $\mathcal{C}^t$ while continuously adapting to both covariate and label shifts, without requiring retraining on additional data. The model adapts to changes in the environment, incorporating new knowledge in real-time as the data distribution and class composition evolve.

Towards the goal of ITTA, we establish a strong benchmark, choosing single-image TTA baselines based on VLMs, ZS-Eval~\cite{clip}, TDA~\cite{tda}, DPE~\cite{dpe}. We integrate these methods with active learning strategies to perform ITTA. We now describe our proposed method SegAssist, a simple plug-in module for active labeling, designed to favour selecting unseen class samples during test time.

\newcommand{\TableMainResultsHMICDD}{
\begin{table*}[t]
\begin{adjustbox}{max width=\linewidth}
\footnotesize
\begin{tabular}{lccccccccccccc}
\toprule
& \multirow{2}{*}{Method} & \multicolumn{2}{c}{ImageNet-R} & \multicolumn{2}{c}{ImageNet-A} & \multicolumn{2}{c}{DN-Clipart} & \multicolumn{2}{c}{DN-Painting} & \multicolumn{2}{c}{DN-Sketch} & \multicolumn{2}{c}{DN-Real}\\
 \cmidrule(r){3-4} \cmidrule(r){5-6} \cmidrule(r){7-8} \cmidrule(r){9-10} \cmidrule(r){11-12} \cmidrule(r){13-14}
&  & HM$\uparrow$ & ICDD $\downarrow$ & HM$\uparrow$ & ICDD $\downarrow$  &   HM$\uparrow$ & ICDD $\downarrow$ & HM$\uparrow$ & ICDD $\downarrow$ & HM$\uparrow$ & ICDD $\downarrow$  &   HM$\uparrow$ & ICDD $\downarrow$  \\ \midrule
\multirow{5}{*}{\rotatebox{90}{ZS-Eval}}  & Random    & 49.57 & 0.2590 &          23.60 &         0.4533  & 29.31 & 0.4496 & 39.41 & 0.3804 & 35.94 & 0.4021 & 57.59 & 0.2988\\
                                        ~ & MSP       & 62.92 & 0.1721 & \textbf{24.59} & 0.4189 & \textbf{40.44} & \textbf{0.3640} & 48.24 & 0.3091 & 47.13 & 0.3264 & 75.86 & 0.1185\\
                                        ~ & Entropy   & 59.89 & 0.2199 &          21.11 &           0.4383 & 40.12 & 0.3688 & 47.06 & 0.3566 & 44.22 & 0.3585 & 66.05 & 0.2017\\
                                        ~ & Margin    & 60.84 & 0.1638 &          20.06 &           0.4579 & 37.34 & 0.3909 & 45.01 & 0.3586 & 42.34 & 0.3647 & 74.25 & 0.1319\\
     \cmidrule(r){2-14}
                                        ~ & SegAssist & \textbf{64.05} & \textbf{0.1601} & 23.94 & \textbf{0.4147} & 40.40 & 0.3660 & \textbf{49.95} & \textbf{0.3028} & \textbf{48.54} & \textbf{0.3066} & \textbf{75.93} & \textbf{0.1165}\\
\midrule
\multirow{5}{*}{\rotatebox{90}{TDA}}      & Random      & 53.33 & 0.2590 & 16.58 & 0.4433 & 30.19 & 0.4496 & 44.20 & 0.3804 & 38.32 & 0.4021 & 59.61 & 0.2988\\
                                        ~ & MSP         & 70.38 & 0.1325 & 19.27 & 0.4393 & 45.30 & 0.3587 & 49.20 & 0.3518 & \textbf{50.47} & \textbf{0.3193} & 75.46 & 0.1325\\
                                        ~ & Entropy     & 64.06 & 0.1952 & 20.02 & 0.4495 & 41.20 & 0.3632 & 48.49 & 0.3584 & 43.26 & 0.3838 & 63.49 & 0.2447\\
                                        ~ & Margin      & 64.85 & 0.1647 & \textbf{21.46} & 0.4427 & 44.31 & 0.3703 & 51.47 & 0.3256 & 46.04 & 0.3520 & 77.86 & 0.1183\\
 \cmidrule(r){2-14}
                                        ~ & SegAssist   & \textbf{70.48} & \textbf{0.1307} & 19.27 & \textbf{0.4393} & \textbf{45.52} &\textbf{ 0.3536} & \textbf{52.25} & \textbf{0.3238} & 49.84 & 0.3316 & \textbf{75.91} & \textbf{0.1284}\\
\midrule
\multirow{5}{*}{\rotatebox{90}{DPE}}      & Random      & 54.90 & 0.2590 & 28.52 & 0.4621 & 31.39 & 0.4496 & 46.31 & 0.3804 & 40.08 & 0.5689 & 60.46 & 0.2988\\
                                        ~ & MSP         & 68.09 & 0.1964 & 24.62 & 0.4718 & 43.64 & \textbf{0.3623} & 54.34 & 0.3259 & 50.04 & 0.5689 & \textbf{73.12} & \textbf{0.1779}\\
                                        ~ & Entropy     & 0.19  & 0.5605 & 4.44  & 0.5788 & 5.21  & 0.5501 & 02.16 & 0.5298 & 00.00 & 0.5689 & 5.28  & 0.5487 \\
                                        ~ & Margin      & 65.47 & 0.1712 & 30.74 & 0.4624 & 42.09 & 0.3764 & 54.05 & 0.3075 & 48.61 & 0.3433 & 71.93 & 0.2074\\
     \cmidrule(r){2-14}
                                        ~ & SegAssist   & \textbf{69.39} & \textbf{0.1638} & \textbf{33.42} & \textbf{0.4598} & \textbf{44.06} & 0.3690 & \textbf{55.70} & \textbf{0.3037} & \textbf{50.28} & \textbf{0.3186} & 73.04 & 0.1810\\
\bottomrule
\end{tabular}
\end{adjustbox}
\caption{ITTA Results: ImageNet-R, ImageNet-A, DN-Clipart, Painting, Sketch, Infograph}
\label{tab:main-results}
\end{table*}
}

\newcommand{\tableVaryingBudget}{
\begin{table}[h]
\centering
% \footnotesize
\begin{tabular}{@{}cccccc@{}}
    \toprule
    \multirow{2}{*}{Budget} & \multirow{2}{*}{Method} & \multicolumn{2}{c}{DN-Painting} & \multicolumn{2}{c}{ImageNet-R} \\
    \cmidrule(l){3-4} \cmidrule(l){5-6}
                            &                         & HM$\uparrow$ & ICDD $\downarrow$ & HM$\uparrow$ & ICDD $\downarrow$          \\
    \midrule
    \multirow{2}{*}{0.5\%}  & ZSEval$^*$   & 62.00   &  50.32      & 52.12         & 26.21         \\
                            & +SegAssist               & \textbf{63.13}          & \textbf{50.28}         & \textbf{52.58}         & \textbf{25.54}         \\
    \midrule
    \multirow{2}{*}{1.0\%}  & ZSEval$^*$   &  70.59        &  64.38        & 62.92         & 17.21         \\
                            & +SegAssist               & \textbf{71.21} & \textbf{65.12}         & \textbf{64.05 }        & \textbf{16.01}         \\
    \midrule
    \multirow{2}{*}{1.5\%}  & ZSEval$^*$   & 73.82         & 32.56       & 68.02         & 13.55         \\
                            & +SegAssist               & \textbf{75.98}          & \textbf{29.93}         & \textbf{68.98}         & \textbf{12.02}         \\
    % \midrule
    % \multirow{2}{*}{2.0\%}  & MSP                     & 57.49          & 24.38         & 69.94         & 11.04         \\
    %                         & SegAssist               & \textbf{57.68}          & \textbf{23.77}         & \textbf{69.97}         & \textbf{11.01}         \\
    \bottomrule
\end{tabular}
\caption{Performance comparison for different budget rates $r$.}\
\label{tab:varying-budget}
\end{table}
}

\newcommand{\tablePluginModule}{
\begin{table}[t]
\centering
% \footnotesize
\begin{tabular*}{\linewidth}{@{\extracolsep{\fill}}ccccc@{}}
\toprule
\multirow{2}{*}{Method}  & \multicolumn{2}{c}{DN-Painting} & \multicolumn{2}{c}{ImageNet-R} \\
\cmidrule(l){2-3} \cmidrule(l){4-5}
                         & HM$\uparrow$ & ICDD $\downarrow$           & HM$\uparrow$ & ICDD $\downarrow$           \\
\midrule
MSP                      & 70.59          &  64.38        & 62.92         & 17.21         \\
SegAssist                & \textbf{71.21} & \textbf{65.12}& \textbf{64.05}& \textbf{16.01}         \\
\midrule
Entropy                  &           &         & 59.88         & 21.99         \\
SegAssist$^*$               & \textbf{} & \textbf{}& \textbf{60.21}& \textbf{21.66}         \\
% \midrule
% Margin                   & 45.01          & 35.86         & 60.84         & 16.37         \\
% SegAssist$^*$               & \textbf{46.26} &\textbf{35.21} &\textbf{62.38} & \textbf{14.27}    \\
\bottomrule
\end{tabular}
\caption{SegAssist with different AL methods for ZS-Eval. SegAssist$^*$ refers to our module used with Entropy and Margin for uncertain sample selection instead of default AL method MSP.}
\label{tab:segassist-acl}
\end{table}
}

\newcommand{\tableVaryingThreshold}{
\begin{table}[h]
    \begin{adjustbox}{max width=\linewidth}
    \centering
\begin{tabular}{@{}cccccc@{}}
    \toprule
    \multirow{2}{*}{Threshold} & \multirow{2}{*}{Method} & \multicolumn{2}{c}{DN-Painting} & \multicolumn{2}{c}{ImageNet-R} \\
    \cmidrule(l){3-4} \cmidrule(l){5-6}
                            &                         & HM$\uparrow$ & ICDD $\downarrow$ & HM$\uparrow$ & ICDD $\downarrow$          \\
    \midrule
    \multirow{2}{*}{0.1}       & ZSEval$^*$    & 59.92           &  49.64        & 52.14         & 29.39         \\
                               & +SegAssist               & \textbf{60.82} & \textbf{48.35}& \textbf{52.14}& \textbf{29.39}         \\
    \midrule
    \multirow{2}{*}{0.2}       & ZSEval$^*$                      &  70.59        & 39.14         & 62.92         & 17.21         \\
                               & +SegAssist               & \textbf{71.21} & \textbf{38.56} & \textbf{64.05}& \textbf{16.01}         \\
    \midrule
    \multirow{2}{*}{0.3}       & ZSEval$^*$      & 66.83          &  45.48        & 62.97         & 13.47         \\
                               & +SegAssist               & \textbf{69.18} & \textbf{43.12} & \textbf{63.91}& \textbf{13.25}         \\
    % \midrule
    % \multirow{2}{*}{0.4}       & MSP                     & 47.20          & 34.27         & 63.83         & 14.76         \\
    %                            & SegAssist               & \textbf{47.59} & \textbf{32.68}& \textbf{64.92}& \textbf{12.10}         \\
    \bottomrule
\end{tabular}
    \end{adjustbox}
\caption{Varying threshold for active sample selection.}
\label{tab:varying-threshold}
\end{table}
}

\newcommand{\tableVaryingRatio}{
\begin{table}[h]
    \centering
    % \begin{adjustbox}{max width=\linewidth}
    %     \begin{tabular*}{\linewidth}{@{\extracolsep{\fill}}ccccccc@{}}
    %     \toprule
    %     \multirow{2}{*}{TTA } & \multirow{2}{*}{AL} & \multicolumn{4}{c}{Ratio of unknown to known} \\
    %     &                            & 0.1        & 0.25         & 0.5         & 0.75        & 1.0           \\
    %     \midrule
    %     \multirow{2}{*}{ZSEval}      & MSP        & 62.30   & 70.59  & 64.38 & 60.21  & 60.50                    \\
    %     & SegAssist                  & \textbf{70.38}       & \textbf{71.21}        & \textbf{65.12}       & \textbf{60.55}   & \textbf{61.10}     \\
    %     \midrule
    %     \multirow{2}{*}{TDA}        & MSP        & 64.34   & 68.83  & 64.96 & 61.17   &  60.28            \\
    %     & SegAssist                  & \textbf{64.74}       & \textbf{69.20}       & \textbf{66.78}       & \textbf{63.21}   & \textbf{62.35}    \\
    %     \midrule
    %     \multirow{2}{*}{DPE}        & MSP        & 63.32   & 71.45  & 64.00  & 60.82 & 60.31   \\
    %     & SegAssist                  & \textbf{66.24}      & \textbf{72.40}       & \textbf{64.56}       & \textbf{61.34}  & \textbf{61.20}   \\
    %     \bottomrule
    %     \end{tabular*}
    % \end{adjustbox}
    \begin{adjustbox}{max width=\linewidth}
        \begin{tabular*}{\linewidth}{@{\extracolsep{\fill}}cccccc@{}}
        \toprule
        \multirow{2}{*}{Method } & \multicolumn{5}{c}{Ratio of unknown to known} \\
            & 0.1        & 0.25         & 0.5         & 0.75        & 1.0           \\
        \midrule
        ZSEval$^*$        & 62.30   & 70.59  & 64.38 & 60.21  & 60.50                    \\
        +SegAssist                  & \textbf{70.38}       & \textbf{71.21}        & \textbf{65.12}       & \textbf{60.55}   & \textbf{61.10}     \\
        \midrule
        TDA$^*$    & 64.34   & 68.83  & 64.96 & 61.17   &  60.28            \\
        +SegAssist                  & \textbf{64.74}       & \textbf{69.20}       & \textbf{66.78}       & \textbf{63.21}   & \textbf{62.35}    \\
        \midrule
        DPE$^*$        & 63.32   & 71.45  & 64.00  & 60.82 & 60.31   \\
        +SegAssist                  & \textbf{66.24}      & \textbf{72.40}       & \textbf{64.56}       & \textbf{61.34}  & \textbf{61.20}   \\
        \bottomrule
        \end{tabular*}
    \end{adjustbox}
    \caption{\color{green}Varying ratio of unknown to known classes.}
    \label{tab:varying-ratio}
\end{table}
}

\newcommand{\tableAnalysis}{
\begin{table*}[t]
% \centering
% \caption{Analysis of different factors affecting performance.}
\begin{threeparttable}
\begin{adjustbox}{max width=\linewidth}
\begin{tabular}{c c c}
    % First subtable
    \begin{minipage}{0.32\linewidth}
        \centering
        \subcaption{Varying budget rates $r$ (HM $\uparrow$ / ICDD $\downarrow$).}
        \label{sub-tab:varying-budget}
        \begin{adjustbox}{max width=\linewidth}
        \begin{tabular}{@{}cccc@{}}
            \toprule
            Budget & Method & DN-Painting & ImageNet-R \\
            \midrule
            \multirow{2}{*}{0.5\%}  
                & ZSEval$^*$   & 62.00 / 50.32  & 52.12 / 26.21  \\
                & +SegAssist   & \textbf{63.13} / \textbf{50.28}  & \textbf{52.58} / \textbf{25.54}  \\
            \midrule
            \multirow{2}{*}{1.0\%}  
                & ZSEval$^*$   & 70.59 / 39.14  & 62.92 / 17.21  \\
                & +SegAssist   & \textbf{71.21} / \textbf{38.56}  & \textbf{64.05} / \textbf{16.01}  \\
            \midrule
            \multirow{2}{*}{1.5\%}  
                & ZSEval$^*$   & 73.82 / 32.56  & 68.02 / 13.55  \\
                & +SegAssist   & \textbf{75.98} / \textbf{29.93}  & \textbf{68.98} / \textbf{12.02}  \\
            \bottomrule
        \end{tabular}
        \end{adjustbox}
    \end{minipage} &
    
    % Second subtable
    \begin{minipage}{0.32\linewidth}
        \centering
        \subcaption{Varying threshold (HM $\uparrow$ / ICDD $\downarrow$).}
        \label{sub-tab:varying-threshold}
        \begin{adjustbox}{max width=\linewidth}
        \begin{tabular}{@{}cccc@{}}
            \toprule
            Thresh. & Method & DN-Painting & ImageNet-R \\
            \midrule
            \multirow{2}{*}{0.1}  
                & ZSEval$^*$   & 59.92 / 49.64  & 52.14 / 29.39  \\
                & +SegAssist   & \textbf{60.82} / \textbf{48.35}  & \textbf{52.14} / \textbf{29.39}  \\
            \midrule
            \multirow{2}{*}{0.2}  
                & ZSEval$^*$   & 70.59 / 39.14  & 62.92 / 17.21  \\
                & +SegAssist   & \textbf{71.21} / \textbf{38.56}  & \textbf{64.05} / \textbf{16.01}  \\
            \midrule
            \multirow{2}{*}{0.3}  
                & ZSEval$^*$   & 66.83 / 45.48  & 62.97 / 13.47  \\
                & +SegAssist   & \textbf{69.18} / \textbf{43.12}  & \textbf{63.91} / \textbf{13.25}  \\
            \bottomrule
        \end{tabular}
        \end{adjustbox}
    \end{minipage} &
    
    % Third subtable
    \begin{minipage}{0.32\linewidth}
        \centering
        \subcaption{Varying ratio of unseen to seen (HM$\uparrow$).}
        \label{sub-tab:varying-ratio}
        \begin{adjustbox}{max width=\linewidth}
        \begin{tabular}{@{}cccccc@{}}
            \toprule
            Method  & 0.1  & 0.25  & 0.5  & 0.75  & 1.0  \\
            \midrule
            ZSEval$^*$        & 62.30  & 70.59  & 64.38  & 60.21  & 60.50  \\
            +SegAssist        & \textbf{70.38} & \textbf{71.21} & \textbf{65.12} & \textbf{60.55} & \textbf{61.10} \\
            \midrule
            TDA$^*$           & 64.34  & 68.83  & 64.96  & 61.17  & 60.28  \\
            +SegAssist        & \textbf{64.74} & \textbf{69.20} & \textbf{66.78} & \textbf{63.21} & \textbf{62.35} \\
            \midrule
            DPE$^*$           & 63.32  & 71.45  & 64.00  & 60.82  & 60.31  \\
            +SegAssist        & \textbf{66.24} & \textbf{72.40} & \textbf{64.56} & \textbf{61.34} & \textbf{61.20} \\
            \bottomrule
        \end{tabular}
        \end{adjustbox}
    \end{minipage} 
\end{tabular}
\end{adjustbox}
\caption{Performance of SegAssist across different ITTA scenarios. $^*$ refers to using MSP as AL method with the TTA method. }
\end{threeparttable}
\label{tab:combined-analysis}
\vspace{-13pt}
\end{table*}
}

\newcommand{\figureMainSegAssist}{
\begin{figure}[t]
           \centering
           \begin{adjustbox}{max width = \linewidth}
           \includegraphics{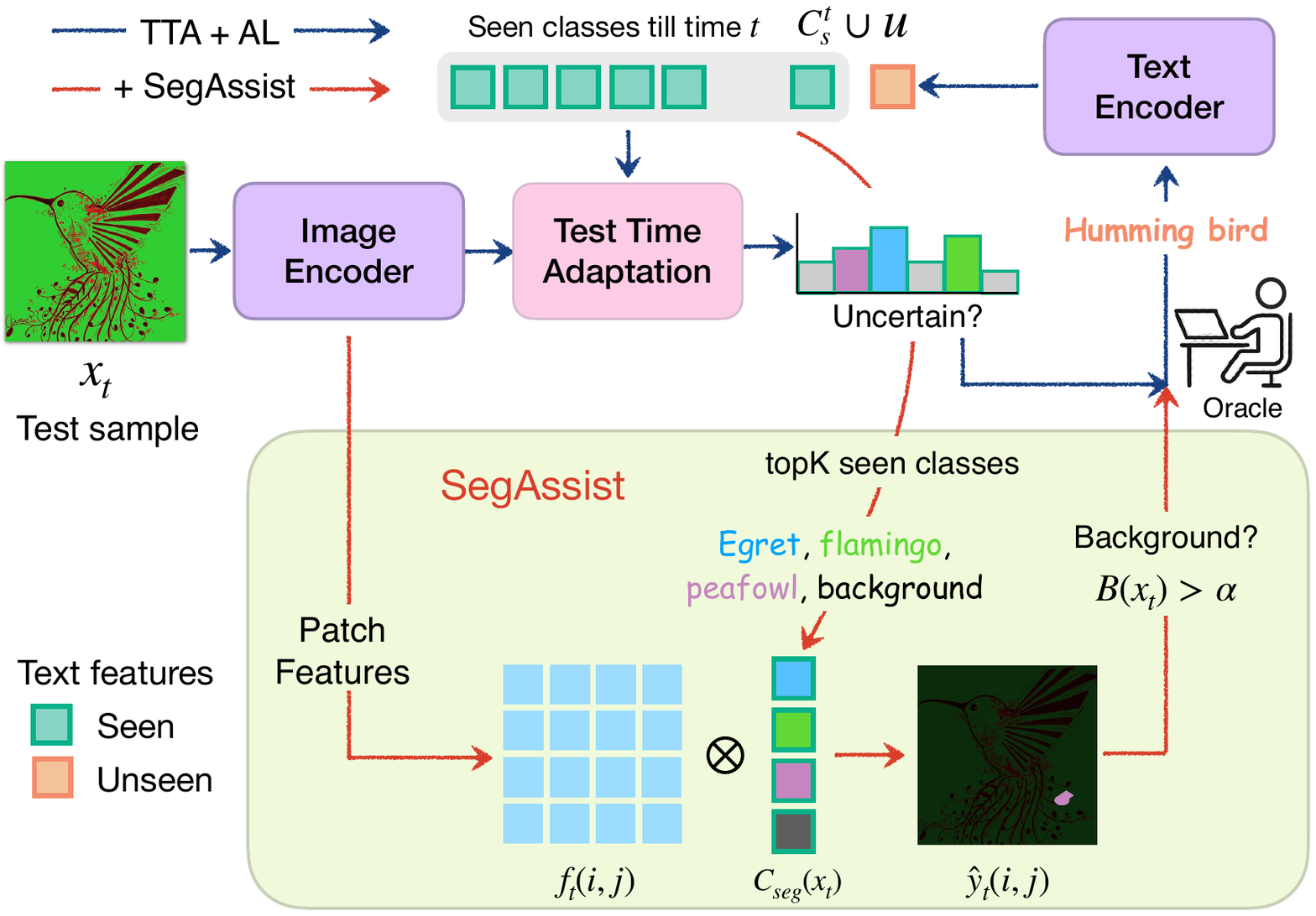}
           \end{adjustbox}
           \caption{SegAssist is a plug-in module that can be integrated with existing TTA and active labeling methods. Uncertain samples are further filtered by performing pixel wise segmentation considering topK classes with “background” class. A sample is selected for labeling by the oracle only if the pixels are predominantly classified as “background”, prioritizing unseen class samples.}
           \label{fig:seg-assist}
           \vspace{-17pt}
\end{figure}
}

\newcommand{\figureIDCC}{
\begin{figure}[t]
           \centering
           \begin{adjustbox}{max width = \linewidth}
           \includegraphics{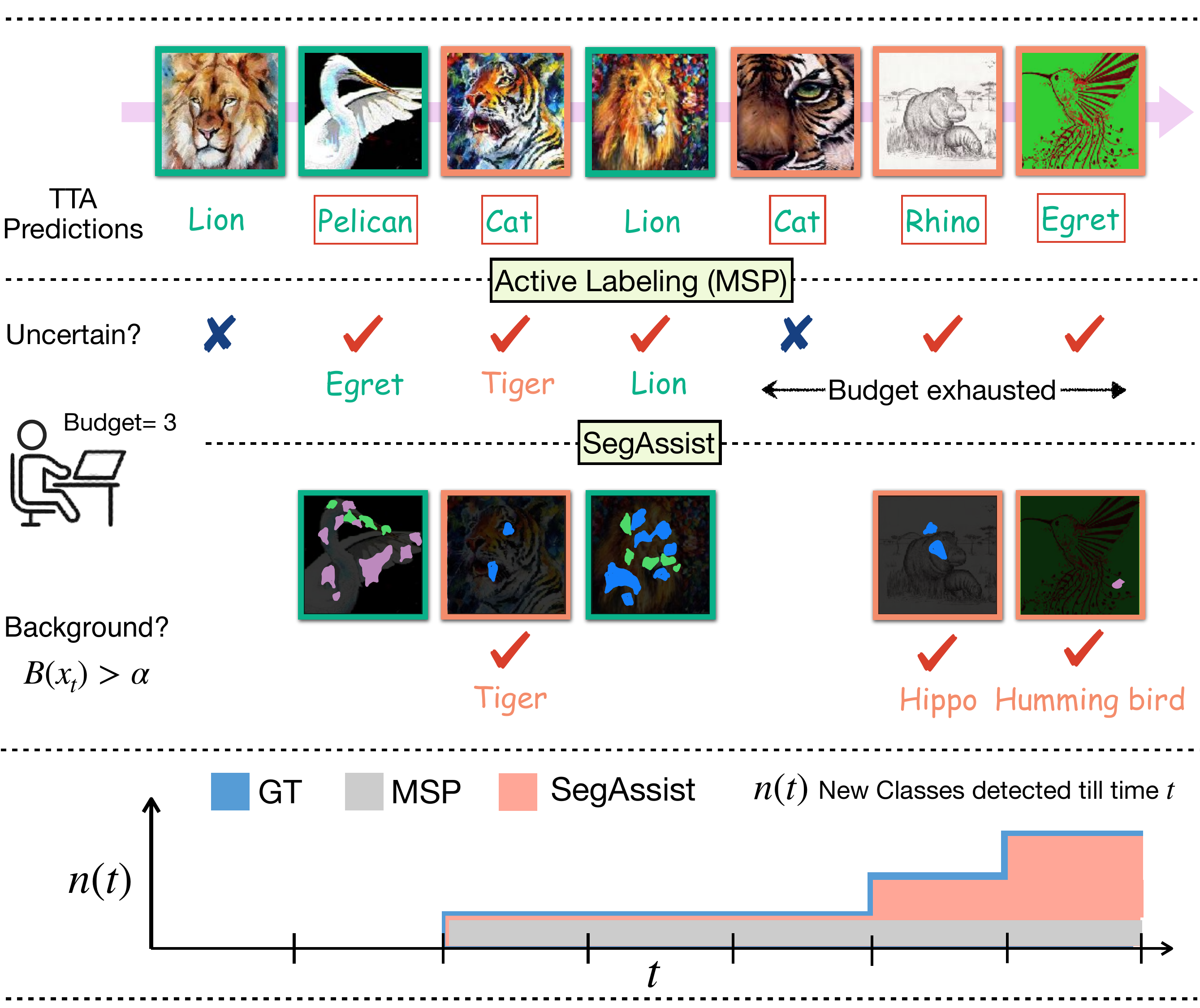}
           \end{adjustbox}
           \caption{Given a budget of 3 samples, active selection based on MSP selects 3 samples, of which two are seen and only one is unseen. The budget being exhausted, the model cannot recognize unseen classes coming later in time. Incorporating SegAssist favors unseen sample selection, hence effectively using the budget. The plot below shows that SegAssist detects all three new classes as they are introduced while MSP detects only one class.}
           \label{fig:seg-atta}
           \vspace{-15pt}
\end{figure}
}

\newcommand{\mainResultsHM}{
\begin{table}[t]
\begin{adjustbox}{max width=\linewidth}
\centering
\footnotesize
% \begin{tabular}{ccccccc}
\begin{tabular*}{\linewidth}{@{\extracolsep{\fill}}cccccc@{}}

\toprule
% & \multirow{2}{*}{Method} & \multicolumn{2}{c}{ImageNet-R} & \multicolumn{2}{c}{ImageNet-A} & \multicolumn{2}{c}{DN-Clipart} & \multicolumn{2}{c}{DN-Painting} & \multicolumn{2}{c}{DN-Sketch} & \multicolumn{2}{c}{DN-Real}\\
 % \cmidrule(r){3-4} \cmidrule(r){5-6} \cmidrule(r){7-8} \cmidrule(r){9-10} \cmidrule(r){11-12} \cmidrule(r){13-14}
% &  & HM$\uparrow$ & ICDD $\downarrow$ & HM$\uparrow$ & ICDD $\downarrow$  &   HM$\uparrow$ & ICDD $\downarrow$ & HM$\uparrow$ & ICDD $\downarrow$ & HM$\uparrow$ & ICDD $\downarrow$  &   HM$\uparrow$ & ICDD $\downarrow$  \\ 
& Method & IN-R & IN-A & Clipart & Painting\\
\midrule

\multirow{4}{*}{\rotatebox{90}{ZS-Eval~\cite{clip}}}  & Random    & 49.57 & 23.60 & 22.04	& 32.54	 \\
                                                    ~ & Entropy   & 59.89 & 21.11 & \underline{48.25} &	64.34 \\
                                                    ~ & MSP       & \underline{62.92} & \textbf{24.59} & 47.65 &	\underline{70.59}\\
     \cmidrule(r){2-6}
                                        ~ & SegAssist & \textbf{64.05} & \underline{23.94} & \textbf{49.08} & \textbf{71.21} \\
\midrule
\multirow{4}{*}{\rotatebox{90}{TDA~\cite{tda}}}      & Random       & 53.33 & 16.58 & 22.54 & 32.66 \\
                                                    ~ & Entropy     & 64.06 & 18.78 & 58.79 & 67.27 \\
                                                    ~ & MSP         & \underline{70.38} & \underline{19.27} & \underline{58.99} &	\underline{68.83} \\
     \cmidrule(r){2-6}
                                             ~ & SegAssist   & \textbf{70.48} & \textbf{20.02} & \textbf{60.25} &	\textbf{69.20}\\
\midrule
\multirow{4}{*}{\rotatebox{90}{DPE~\cite{dpe}}}      & Random      & 54.90 & \underline{28.52} & 23.36 &	33.49 \\
                                                   ~ & Entropy     & 0.19  & 4.44  &  15.50 &	00.00 \\
                                                   ~ & MSP         & \underline{68.09} & 24.62 & \underline{65.35} &	\underline{71.45}\\
     \cmidrule(r){2-6}
                                                   ~ & SegAssist   & \textbf{69.39} & \textbf{33.42} & \textbf{65.49} &	\textbf{72.40} \\
\bottomrule
\end{tabular*}
\end{adjustbox}
\caption{Comparison of Harmonic Mean (higher is better) of seen and unseen class accuracy across different Active labeling and TTA Methods employed in the ITTA framework.}
\label{tab:main-results-hm}
\vspace{-10pt}
\end{table}
}

\newcommand{\mainResultsICDD}{
\begin{table}[t]
\begin{adjustbox}{max width=\linewidth}
\centering
\footnotesize
% \begin{tabular}{ccccccc}
\begin{tabular*}{\linewidth}{@{\extracolsep{\fill}}cccccc@{}}

\toprule
% & \multirow{2}{*}{Method} & \multicolumn{2}{c}{ImageNet-R} & \multicolumn{2}{c}{ImageNet-A} & \multicolumn{2}{c}{DN-Clipart} & \multicolumn{2}{c}{DN-Painting} & \multicolumn{2}{c}{DN-Sketch} & \multicolumn{2}{c}{DN-Real}\\
 % \cmidrule(r){3-4} \cmidrule(r){5-6} \cmidrule(r){7-8} \cmidrule(r){9-10} \cmidrule(r){11-12} \cmidrule(r){13-14}
% &  & HM$\uparrow$ & ICDD $\downarrow$ & HM$\uparrow$ & ICDD $\downarrow$  &   HM$\uparrow$ & ICDD $\downarrow$ & HM$\uparrow$ & ICDD $\downarrow$ & HM$\uparrow$ & ICDD $\downarrow$  &   HM$\uparrow$ & ICDD $\downarrow$  \\ 
& Method & IN-R & IN-A & Clipart & Painting\\
\midrule

\multirow{4}{*}{\rotatebox{90}{ZS-Eval~\cite{clip}}}  & Random    & 25.90 & 45.33 & 75.30 & 68.41 \\
                                                    ~ & Entropy   & 21.99 & 43.83 & \underline{61.88} & 43.89 \\
                                                    ~ & MSP       & \underline{17.21} & \textbf{41.47} & 62.04 & \underline{39.14} \\
     \cmidrule(r){2-6}
                                        ~ & SegAssist & \textbf{16.01} & \underline{41.89} & \textbf{61.20} & \textbf{38.56} \\
\midrule
\multirow{4}{*}{\rotatebox{90}{TDA~\cite{tda}}}      & Random       &  25.90 & 45.33 & 75.30 & 68.41\\
                                                    ~ & Entropy     &  19.52 & 44.95 & 49.10 & 46.08 \\
                                                    ~ & MSP         & \underline{13.25} & \underline{43.93} & \underline{50.68} &	\textbf{39.31} \\
     \cmidrule(r){2-6}
                                             ~ & SegAssist   & \textbf{13.07} & \textbf{43.93} & \textbf{48.96} &	\underline{41.27}\\
\midrule
\multirow{4}{*}{\rotatebox{90}{DPE~\cite{dpe}}}      & Random      & 25.90 & \underline{45.33} & 75.30 & 68.41  \\
                                                   ~ & Entropy     & 56.05 & 57.88 & 80.27 & 81.81 \\
                                                   ~ & MSP         & \underline{19.64} & 47.18  & \underline{49.00} &	\underline{34.70}\\
     \cmidrule(r){2-6}
                                                   ~ & SegAssist   & \textbf{16.38} & \textbf{44.98} & \textbf{49.00} &	\textbf{33.12} \\
\bottomrule
\end{tabular*}
\end{adjustbox}
\caption{Comparison of ICDD (lower is better) across different Active labeling and TTA Methods  in the ITTA framework.}
\label{tab:main-results-icdd}
\vspace{-10pt}
\end{table}
}

\section{Proposed Framework}
\label{sec:segmentation-atta}

In Incremental Test-Time Adaptation (ITTA), selecting informative samples for active labeling is critical in budget-constrained scenarios where querying each uncertain sample may not be feasible. If a test sample’s confidence score is below a predefined threshold, the oracle is queried for active labeling. However, with limited budget, it becomes crucial to refine this pool further, to select samples that are likely to belong to unseen classes. 
% A more refined selection would maximize the informativeness of queried samples, specifically by increasing the likelihood of identifying unknown class instances. 
Towards this goal, we propose a novel module to further refine the samples obtained using existing active labeling strategies like confidence, entropy based uncertainty scores, etc.

\subsection{Can CLIP scores identify New Classes?} 
Analyzing the Maximum Softmax Probability (MSP) scores obtained by matching CLIP’s global image features, from Fig.~\ref{fig:clip-scores}, we observe that the confidence scores for known and unknown class samples exhibit substantial overlap.
This makes it challenging to reliably distinguish unseen from seen classes based on scores like MSP alone, potentially leading to inefficient use of the available budget $\mathcal{B}_t$ as seen class samples may be selected more frequently than unseen class samples. Here, we explore whether fine-grained, local image features can be leveraged to improve the ability to discriminate between seen and unseen classes.

\begin{figure}[t]
    \centering
    % Place the common legend above the subfigures
    \pgfplotslegendfromname{commonlegend}
    \vspace{0.15cm} % Adjust vertical space between legend and subfigures if needed
    \begin{subfigure}[b]{0.49\linewidth}
        \centering
        \begin{adjustbox}{max width=\linewidth}
            \begin{tikzpicture}
                \begin{axis}[
                    xlabel={\Huge $s_t$},
                    xmin=0.,
                    xmax=1.0,
                    ymin=0.,
                    ymax=970,
                    axis background/.style={fill=gray!0},
                    tick label style={font=\Large},
                    xtick style={draw=none},
                    ymajorgrids=true,
                    grid style={dashed, gray!30},
                    legend to name=commonlegend, % Define the legend name
                    legend columns=2,
                    legend style={/tikz/every even column/.append style={column sep=10pt}}
                ]
                    \addplot[
                        seen!10!white,
                        fill=seen,
                        hist,
                        hist/bins=100,
                        opacity=0.7,
                        area legend,
                    ] table[
                        y index=1,col sep=comma
                    ] {data/scores/painting_known_all.csv};

                    \addplot[
                        unseen!10!white,
                        fill=unseen,
                        hist,
                        hist/bins=100,
                        opacity=0.7,
                        area legend,
                    ] table[
                        y index=1,col sep=comma
                    ] {data/scores/painting_unknown_all.csv};
                    \legend{Seen classes, Unseen classes}
                \end{axis}
            \end{tikzpicture}
        \end{adjustbox}
        \caption{\small MSP scores for Painting}
        \label{fig:clip-scores}
    \end{subfigure}
    \hfill
    \begin{subfigure}[b]{0.5\linewidth}
        \centering
        \begin{adjustbox}{max width=\linewidth}
            \begin{tikzpicture}
                \begin{axis}[
                    ybar=0pt,
                    bar width=27pt,
                    ymin=75,
                    ymax=100,
                    symbolic x coords={Painting, ImageNetR},
                    xtick=data,
                    xtick style={draw=none},
                    tick label style={font=\Large},
                    ytick = {80,85,90,95},
                    axis background/.style={fill=gray!0},
                    ylabel style={yshift=0.02cm},
                    xlabel style={yshift=0.25cm},
                    enlarge x limits=0.4,
                    nodes near coords,
                    every node near coord/.append style={font=\large},
                    nodes near coords align={vertical},
                    ymajorgrids=true,
                    grid style={dashed, gray!30},
                ]
                    % Known data
                    \addplot+[ybar, seen, fill=seen!70!white] plot coordinates {(Painting, 84.38) (ImageNetR, 93.55)};
                    % Unknown data
                    \addplot+[ybar,unseen, fill=unseen!70!white] plot coordinates {(Painting, 89.35) (ImageNetR, 97.27)};
                \end{axis}
            \end{tikzpicture}
        \end{adjustbox}
        \caption{\% of background images.}
        \label{fig:bkg-statistics}
    \end{subfigure}

    \caption{(a) The histogram of MSP scores obtained based on global features overlap significantly making it very challenging to distinguish between seen and unseen classes. (b) Among the uncertain samples, segmentation maps obtained from local features can help distinguish them, as more unseen class samples are characterized as background compared to seen class samples.}
    \vspace{-15pt}
\end{figure}

To investigate this hypothesis, we perform pixel-wise segmentation for all uncertain samples, aiming to identify distinctive patterns between samples from seen vs. unseen classes. Using CLIP, we perform dense predictions on the image~\cite{maskclip} by matching the patch features with the top-K predictions and a “background” class. For seen class samples, we expect the model to correctly classify the pixels into one of the top-K classes predicted, while samples from unseen classes should primarily register as “background” across most pixels. While this approach may not be perfectly accurate, as foreground pixels from unseen classes could map to related seen classes within the top-5 predictions, it can still aid in distinguishing seen and unseen class samples. Our analysis from Fig.~\ref{fig:bkg-statistics} reveals that, among the uncertain samples, the number of background images (majority of pixels classified as background) is more for unseen class samples compared to uncertain seen class samples.
% in the segmentation map can serve as a key signal for distinguishing between known and unknown class samples. For samples in the unknown class, the majority of pixels often fall under the “background” category, highlighting the model's struggle to map these features to known categories.

\figureMainSegAssist

Based on this insight, we propose SegAssist to effectively select samples for active labeling. Only uncertain samples with over 95\% of pixels classified as “background” are selected. 
% These uncertain samples can belong to both known and unknown classes as shown in Fig.~\ref{fig:seg-atta}. 
% However, this addition step of filtering is designed to increase the probability of selecting unseen class samples.
%Operating within highly constrained active learning budget, SegAssist aims to further increase the probability of selecting unknown class samples for active querying in the ITTA framework. 
Thus, SegAssist aims to refine the sample selection process, enhancing the budget utilization by prioritizing those test samples that are most likely to represent new, unseen classes. 
{\em SegAssist is a simple, yet effective plug-in module, requiring no additional model training. With no computation overhead, it can be seamlessly integrated with any existing TTA and active labeling method to adapt them to the realistic ITTA setting. }
% We now describe the module in detail.

\subsection{SegAssist for Active Sample Selection}
The first step of active sample selection is to identify uncertain samples. 
Given an incoming sample \( x_t \in \mathcal{X} \) from the test stream, we can utilize any standard measure like confidence. entropy, etc. to filter the initial set of uncertain samples. 
By default, we use the Maximum Softmax Probability (MSP) calculated as
$s_t = \max_{c} p_{c}(x_t)$ to identify uncertain samples. Here, $p_{c}(x_t)$ is the softmax score predicted for class $c$. We consider $x_t$ as {\em uncertain} if:
$s_t < \tau$,
where $\tau$ is the uncertainty threshold.
Further, we refine this selection through the proposed SegAssist module, which consists of two key steps: (i) Generating pixel-wise segmentation maps, (ii) 
Selecting samples for active learning based on background pixel proportion, as shown in Fig.~\ref{fig:seg-assist}. 

% \noindent\textbf{Identifying Uncertain Samples: }
% Given an incoming sample \( x_t \in \mathcal{X} \) from the test stream, we can utilize any standard measure like confidence. entropy, etc. to filter the initial set uncertain samples. 
% For majority of our experiments, we have used prediction confidence (\( f(x_t) \)) produced by the CLIP model for the current set of known classes \( C_s \). 
% The confidence score for $x_t$ is computed based on global features.
% We consider $x_t$ as {\em uncertain} if:
% $\text{Conf}(x_t) < \tau$,
% where \( \tau \) is a threshold confidence level. 

\noindent\textbf{(i) Pixel-Wise Segmentation of Uncertain Samples.}
For each uncertain sample $x_t$, let $topK(x_t)$ denote the top-K class names as predicted by the model. 
We define the set $C_{\text{seg}}(x_t) = topK(x_t) \cup \{\text{``background"}\}$ as the classes of interest to perform pixel-wise segmentation. 
% We then use \( C_{\text{seg}} \cup \{\text{background}\} \) as the set of labels for pixel-wise segmentation, where each pixel \( p_i \) in the segmentation map \( S(x) \) is assigned a label from \( C_{\text{seg}} \cup \{\text{background}\} \) based on dense predictions from CLIP.
Though global image feature, which is the embedding of the [CLS] token is used for classification in CLIP, we also have access to the patch features, which can be leveraged to get dense predictions. Following~\cite{maskclip, clip-dinoiser}, we obtain the patch features using the value vectors from the last self attention layer followed by the final projection layer.
We $f_t(i,j)$ denote the feature of patch $(i,j)$ of image $x_t$. Its prediction is obtained as 
\begin{equation}
\hat{y}_t(i,j) = \underset{c\in C_{\text{seg}}(x_t)}{\arg \max} \langle f_t(i,j), t_c \rangle
\end{equation}
where $t_c$ is the text embedding of class $c\in C_{seg}(x_t)$. These predictions at the patch level are upsampled using bilinear interpolation to obtain the segmentation map $S(x_t)$.

% ************************
\mainResultsHM

\noindent\textbf{(ii) Background Ratio for Active Sample Selection.}
To assess if an uncertain sample $x_t$ is more likely to belong to an unseen class, we compute the proportion of pixels in \( S(x_t) \) classified as ``background". 
We denote \( |S(x_t)| \) as the total number of pixels in \( S(x_t) \). The background ratio $B(x_t)$ is then computed as:
\begin{equation}
B(x_t) = \frac{1}{|S(x_t)|} \sum_{i=1}^{|S(x_t)|} \mathbb{I}[S_i(x_t) = \text{``background"}],
\end{equation}
where \( \mathbb{I}[\cdot] \) is an indicator function, returning 1 if \( S_i(x_t) = \text{background} \) and 0 otherwise.

%This background ratio serves as a heuristic measure; 
We hypothesize that a high proportion of background pixels indicate that $x_t$ belongs to an unknown class, as samples from known classes are expected to contain foreground predictions that align with seen classes \( C_s \).
%\noindent\textbf{Step 4: Selection for Active Learning.}
Thus, we filter uncertain samples for active labeling based on their background ratio. 
Specifically, we select $x_t$ for querying only if $B(x_t) > \alpha $
where $\alpha$ is a threshold, set to 0.95. This ensures that samples with a significant proportion of pixels classified as ``background"  are prioritized for active labeling.
%enhancing the probability of selecting genuinely novel samples from unknown classes.

% \begin{figure}[t]
%            \centering
%            \begin{adjustbox}{max width = \linewidth}
%            \includegraphics{data/SegAssist-block-diagram.pdf}
%            \end{adjustbox}
%            \caption{ \ToDo{SegAssist Framework.}}
%            \label{fig:seg-assist}
% \end{figure}

% \begin{figure}[t!]
%            \centering
%            \begin{adjustbox}{max width = \linewidth}
%            \includegraphics{data/acl-metric}
%            \end{adjustbox}
%            \caption{ \ToDo{ ICDD}: Classes increase with time; To be detected. AUC(Ground Truth) - AUC(Active learning method) will measure the performance of the active learning method.}
%            \label{fig:acl-metric}
% \end{figure}

In essence, {\em SegAssist refines the selection process in ITTA, leveraging pixel-wise segmentation maps to prioritize samples that are genuinely novel, making the budget utilization and adaptation process more effective}, as also illustrated in Fig.~\ref{fig:seg-assist}. By leveraging local features, SegAssist goes beyond traditional active labeling approaches to improve sample selection in open-world settings, where incorporating new classes dynamically is integral to the performance of the model. It can be integrated with any VLM based TTA framework, with practically no computation overhead. 
{\em As the patch features are already computed, the only additional computation is the comparison of patch features with $C_{\text{seg}}$, thus making it a very light weight module.}

\section{Incremental Class Detection Delay (ICDD)}
\label{sec:metric-icdd}

\figureIDCC

In an open-world, where new classes appear dynamically with time, a key challenge for ITTA approaches is detecting these emerging classes as they are introduced. 
Towards this objective, we design \textit{Incremental Class Detection Delay (ICDD)}, a metric to quantify the performance of an ITTA system in identifying new classes in a timely manner.
It compares the rate of new class introductions in the \textit{ground truth sequence} with the rate of class detections by the ITTA framework (bottom plot in Fig.~\ref{fig:seg-atta}), visualized as a normalized Area Under the Curve (AUC) plot. For each normalized time step $t$, we track two cumulative functions:
\begin{itemize}
    \item $n_{\text{gt}}(t)$: the cumulative count of classes introduced till time $t$ in the ground truth sequence.
    \item $n_{\text{det}}(t)$: the cumulative count of classes detected by the ITTA system till time $t$.
\end{itemize}
Both $n_{\text{gt}}(t)$ and $n_{\text{det}}(t)$ are normalized between 0 and 1 with respect to the total number of new classes. Similarly, time $t$ is normalized to the interval \([0,1]\), representing the fraction of the total test stream duration. This yields two cumulative curves spanning the area from (0,0) to (1,1), where $n_{\text{det}}(t)$ lags behind $n_{\text{gt}}(t)$ when there is a delay in the detection of an unseen class.
The ICDD score is defined as the difference in AUC between these two curves
\begin{equation} 
\text{ICDD} = \text{AUC}(n_{\text{gt}}(t)) - \text{AUC}(n_{\text{det}}(t))
\end{equation}

\noindent ICDD measures the delay in detecting new classes relative to their actual introduction in the test stream.
A lower ICDD score (closer to zero) suggests that the ITTA system is highly responsive, detecting new classes almost immediately as they appear. 
Conversely, a higher score indicates a greater delay in identifying unseen classes, thereby leading to a decrease in the performance.
ICDD facilitates comparison of different active learning and adaptation strategies, providing insights into their effectiveness in an open-world.
%This metric is particularly relevant for evaluating ITTA systems under open-world conditions where timely adaptation is essential. By quantifying the gap between class introductions and detections, ICDD facilitates the comparison of different active learning and adaptation strategies, providing insights into their effectiveness in discovering unknown classes and highlighting areas for improvement in the adaptability of ITTA systems.

% \begin{figure}[t!]
%            \centering
%            \begin{adjustbox}{max width = \linewidth}
%            \includegraphics{data/acl-metric}
%            \end{adjustbox}
%            \caption{ \ToDo{ ICDD}: Classes increase with time; To be detected. AUC(Ground Truth) - AUC(Active learning method) will measure the performance of the active learning method.}
%            \label{fig:acl-metric}
% \end{figure}

\mainResultsICDD

\begin{figure*}
    \centering    \includegraphics[width=\linewidth]{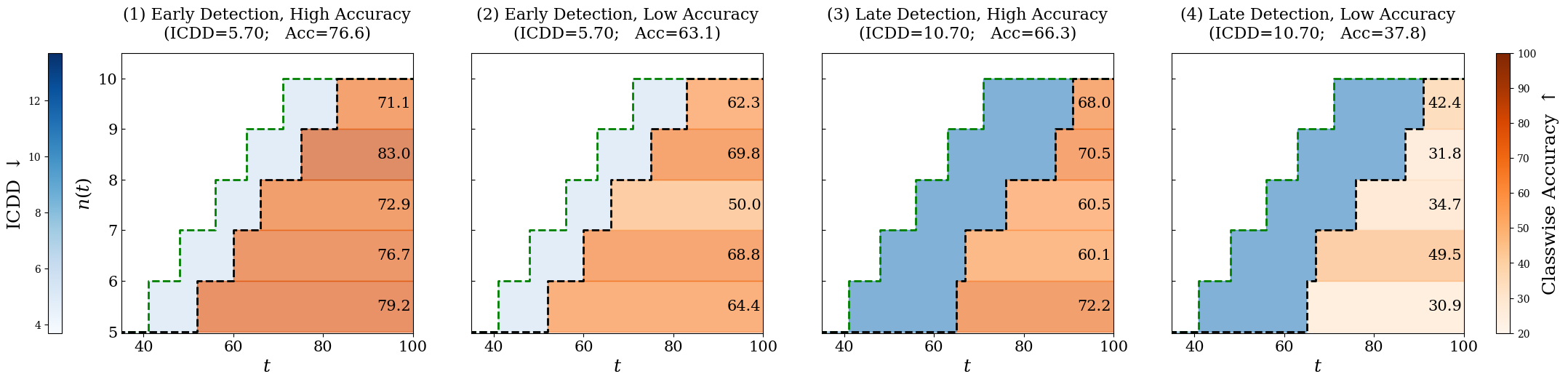}
    \caption{This figure illustrates the impact of early and late detection on the total number of classes recognized over time. The green dashed line represents the ground truth (ideal detection), while the black dashed line shows the detected classes under different scenarios. The shaded blue region quantifies the {\bf ICDD}. Classwise accuracy of each newly detected class is shown at the detected time step.}
    \label{fig:icdd-scenarios}
\vspace{-10pt}
\end{figure*}

\noindent{\bf Complementary nature of ICDD and Accuracy Metrics}

\noindent Here, we construct different ITTA scenarios to illustrate the complementary nature of ICDD and accuracy metrics in evaluating ITTA. 
% This enables us to analyze different detection behaviors by varying the timeliness and correctness of detection.
The performance of ITTA methods are influenced by two key factors:  
{\bf (a) Detection Delay}, where classes are either detected early or late. The extent of detection delay depends on the interaction between the active learning strategy and the test-time adaptation (TTA) method, both of which influence how efficiently the model discovers and integrates new information. To quantify detection delay, we propose the {\em Incremental Class Detection Delay (ICDD)}, which measures the discrepancy between the ground-truth class introduction curve and the detected class curve. A {\em lower} ICDD value indicates that classes are detected in a timely manner, whereas a {\em higher} ICDD signifies significant detection delay.  
{\bf (b) Accuracy of Detected Classes}, where the newly detected unseen classes may be correctly or incorrectly classified. The classification accuracy is determined by the combined effect of the active selection strategy and the chosen TTA method. Even with a similar detection timeline (i.e., comparable ICDD values), different TTA methods can yield varying accuracies. For example, under comparable ICDD values, a method such as TDA~\cite{tda} may achieve higher accuracy than ZS-Eval (Zero-Shot Evaluation), demonstrating the independent role of TTA in improving classification reliability.  

To analyze the interplay between detection delay and classification accuracy, we consider four distinct detection scenarios which we illustrate in Fig.~\ref{fig:icdd-scenarios} and describe below:

\noindent {\bf (1) Early Detection, High Accuracy}: Classes are detected early (low ICDD), often facilitated by effective active selection strategies such as MSP~\cite{msp}. Additionally, high classification accuracy is achieved using strong TTA methods such as TDA~\cite{tda}, making this the ideal scenario.  

\noindent {\bf (2) Early Detection, Low Accuracy}: Here, detection is timely (low ICDD), but classification accuracy remains poor due to weaker baselines such as ZS-Eval, which simply applies zero-shot evaluation of CLIP on newly detected classes without adaptation.  

\noindent {\bf (3) Late Detection, High Accuracy}: The model detects classes with a delay (high ICDD), often the case with simple selection strategies like Random sampling, but achieves high classification accuracy due to stronger TTA methods.  

\noindent {\bf (4) Late Detection, Low Accuracy}: This represents the worst-case scenario, where the model not only detects new classes late (high ICDD) but also struggles with classification, leading to poor overall performance.

\noindent These observations emphasize the necessity of considering both ICDD and classification accuracy as they offer complementary perspectives on evaluating the overall effectiveness of an ITTA method. We discuss such scenarios from our experiments in the Appendix~\ref{app:metrics-discussion}. We report two key metrics: (1) the Harmonic Mean (HM) of accuracy on seen and unseen classes, and (2) ICDD, which together provide a comprehensive assessment of the ability of an ITTA method to both {\em discover} and {\em correctly classify} new classes. Although ICDD captures the timeliness of detection, whereas accuracy evaluates how well the detected classes are classified, we emphasize that we do not attribute ICDD solely to active labeling and HM to TTA performance. Rather, it is crucial to recognize that both metrics reflect the combined impact of both components, reinforcing the need for a holistic approach in ITTA evaluation.

\section{Experimental Evaluation}
\label{sec:baselines}
Here, we describe the experiments performed to evaluate the effectiveness of the proposed framework. 
% First, we describe the baselines used in our work.
% We consider two categories of baselines: (i) Single image TTA methods using VLMs; (ii) Active learning methods for sample selection.

% The former provides a foundation for evaluating the performance of ITTA methods with VLMs, while the later enables us to study the impact of active learning on ITTA in real-time scenarios.
% *******************************
\subsection{Baseline VLM-based Single Image TTA}
\label{sec:baselines-tta}
%\begin{enumerate}
%    \item
The proposed SegAssist active labeling module can be seamlessly used with existing VLM-based TTA methods~\cite{tda, dpe}. We plug-in SegAssist with Zero-Shot Evaluation (ZS-Eval) and state-of-the-art TTA approaches TDA~\cite{tda}, DPE~\cite{dpe} to demonstrate its effectiveness. \\
\textbf{1) ZSEval~\citep{clip}:} A test image $x_{t}$ is classified by matching its image feature $f_t=\mathcal{\phi}_{I}(x_t)$ with text features $\phi_T(c)$ as  $\hat{y}_t = \arg \max_{c \in \mathcal{C}^t} \langle \phi_I(x_t), \phi_T(c) \rangle$, where $C^t$ refers to the classes seen till time $t$ in ITTA scenario.

% , where $c \in \mathcal{C}^t$ and the class prediction is made by identifying the text feature with highest similarity $y_t = \arg \max_{c \in \mathcal{C}^t} \langle \phi_I(x_t), \phi_T(c) \rangle$.

%    For a $C$-class classification problem, the classifier is obtained by prepending a predefined text prompt $\boldsymbol{p}_{T}$="A photo of a", with the class names $\{c_1, c_2,\ldots c_C\}$ to form class specific text inputs $\{\boldsymbol{p}_T, c_i\}$ for $i\in\{1,\ldots C\}$. These texts are then embedded through the text encoder as $\boldsymbol{t}_{i}=\mathcal{F}_{T}(\{\boldsymbol{p}_{T};c_{i}\})$ to get the text classifiers $\{\boldsymbol{t}_1, \boldsymbol{t}_2,\ldots \boldsymbol{t}_C\}$. The class prediction is made by identifying the text feature $\boldsymbol{t}_i$ which has the highest similarity with the image feature $f_t$.
%    \item \textbf{TPT~\cite{tpt}} aims to improve the zero shot generalization ability of CLIP by providing custom adaptable context for each image. This is done by prepending learnable text prompts $\boldsymbol{p}_{T}$ to the class names instead of a predefined text prompt. The text classifiers $\boldsymbol{t}_{i}=\mathcal{F}_{T}(\{\boldsymbol{p}_{T};c_{i}\}), i\in \{1,2,\ldots C\}$ are now a function of these learnable prompts, which are specially adapted for each test image using an entropy minimization objective as  $\arg \min _{\boldsymbol{p}_T} \mathcal{L}_{\text {ent }}$. The entropy is obtained using the average score vector of the filtered augmented views.
    \noindent\textbf{2) TDA~\cite{tda} (CVPR'24)}: They employ a training-free dynamic adapter to perform TTA of VLMs. This is done through a lightweight key-value cache to efficiently refine pseudo labels without backpropagation.

    \noindent\textbf{3) DPE~\cite{dpe} (NeurIPS'24)}: Here, they perform TTA by effectively accumulating task-specific knowledge from both textual and visual modalities through a Dual Prototype Evolving (DPE) approach. They also optimize learnable residuals to align the prototypes from both modalities.
%\end{enumerate}

\tableAnalysis

\subsection{Baseline Active Labeling Methods}
To enable a model deployed in real world to discover new classes, we propose to use Active Labeling (AL) to identify test samples to be queried by an oracle.
% enabling the model to incorporate new classes with time.
% determine and query labels for uncertain test samples. 
%In open-world and incremental learning scenarios, active learning (AL) methods can be used to query labels for uncertain test samples, enabling us to incorporate new classes if present in the test stream. 
Most recent AL techniques require abundant data~\cite{kmeans, activeda, clue, atta} to identify active samples, which may not be suitable for single-sample ITTA scenario which is of our interest.
Consequently, we employ AL methods that enable querying on a per-sample basis, which we now describe.
% allowing for effective querying of individual samples in real-time. Below, we describe the baseline AL techniques used here. 

\noindent\textbf{1. Random:} This is the simplest AL baseline, in which a test sample is randomly selected for querying with probability $r$. Here, $r$ is the predefined budget rate defining the budget $B_t$ as described in Section~\ref{sec:budget-itta}.
% based on a \color{red} predefined budget rate $r$. A sample is selected for active labeling with probability $r$.
% \color{black} which we set as the probability of querying a sample. For each sample, a random number $r$ is generated, and if $  < r $, the sample is queried. 
% This method serves as a baseline to measure the effectiveness of more informed selection criteria.

\noindent\textbf{2. Maximum Softmax Probability (MSP):} This method selects samples based on the MSP score $s_t = \max_{c\in C^t} p_{c}$ predicted by the model. Here, $p_{c}$ is the predicted probability for class $c$. Samples are queried if their MSP score falls below a threshold, indicating low confidence.
%This approach is motivated by the idea that samples with lower confidence are more likely to belong to unknown classes or exhibit characteristics that differ from the model’s training distribution, thus representing informative cases for querying.

\noindent\textbf{3. Entropy:} The uncertainty of model predictions can be measured using the entropy measure $H(x_t) = -\sum_{c\in C^t} p_{c} \log p_{c}$  and is frequently used to identify samples for active labeling. 
% The entropy $H$ of a sample $x_t$ is calculated as $H(x_t) = -\sum_{c\in C^t} p_{c} \log p_{c}$.
% , where $C^t$ is the number of classes known till time $t$ and $ \{p_{c}\}_{c=1}^{C^t} $ are the predicted class probabilities. 
Test samples with entropy above a predefined threshold are selected for querying.
% as higher entropy indicate more uncertainty in the prediction.

% \noindent\textbf{4. Margin:} Margin-based sampling selects samples based on the difference between the two highest predicted probabilities. For a sample $x_t$, let $p_1$ and $p_2$ denote its top two softmax scores. The margin score $ M(x_t)$ is calculated as $M(x_t) = p_1 - p_2$. A smaller margin suggests a high level of ambiguity between the top classes, making it a “confusing” sample for the model to classify. Test samples with margin below a threshold are queried, as they likely contain information useful for disambiguating closely related classes or a potential new class sample.

%We incorporate each of these active learning methods with the baseline TTA methods defined in Section~\ref{sec:baselines-tta} to evaluate their performance of ITTA with limited annotation budget $\mathcal{B}$ rate. Random sampling provides a baseline without any prioritization, while entropy, confidence and margin-based sampling methods allow for more targeted queries based on uncertainty. These baselines provide a foundation for evaluating the effectiveness of more sophisticated active learning approaches in single sample ITTA scenarios.

%\section{Experiments}
%\label{sec:experiments}

% \input{sec/tables}

% \TableMainResultsHMICDD

\subsection{Datasets and Implementation Details}

We use the following large-scale benchmark datasets for evaluation.  
\noindent \textbf{ImageNet-R}~\cite{imagenetr}  is a realistic style transfer dataset encompassing interpretations
of 200 ImageNet classes.
% We split the classes into 160 seen and the other 40 as unseen classes, in the ratio of 80:20.
\textbf{ImageNet-A}~\cite{imageneta} is an adversarial style transfer dataset, containing of 7,500 images belonging to 200 categories. 
% We split the classes into 160 seen and the other 40 as unseen classes, in the ratio of 80:20.
We perform experiments on \textbf{Clipart} and \textbf{Painting} domains from the \textbf{DomainNet}~\cite{domainnet} dataset.
% \textbf{DomainNet}~\cite{domainnet} is a large-scale domain adaptation dataset. We perform experiments on \textbf{Clipart}, \textbf{Painting}, \textbf{Sketch}, and \textbf{Real} domains, each containing 345 classes. 
% We split the classes into 273 seen and the other 72 as unseen classes, in the ratio of 80:20.
% *******************************
%\subsection{Implementation Details}
We split the total number of classes in each dataset into seen and unseen classes, in the ratio of 4:1 to create the test stream. 
% For ImageNet-R and ImageNet-A, we use 160 seen and 40 unseen classes which are introduced in four phases, 10 at a time. For DomainNet, we use 100 seen and 25 unseen classes, which are introduced in five phases, 5 at a time. 
We use a budget rate $r$ as 1\% and the budget $\mathcal{B}_t$ is replenished with 10 samples for every 1000 test samples. We set the active labeling threshold for MSP and Entropy as 0.2 and 0.5 and respectively. 
For SegAssist, we consider the top 5 classes predicted by the model and ``background" to perform pixelwise segmentation. 
We use CLIP ViT-B16~\cite{clip} model for all our experiments. For baseline TTA methods, we use the same hyperparameters used in their works~\cite{tda, dpe}. We run all experiments on an NVIDIA A6000 graphics card.
\color{black}

%\subsection{Evaluation Metrics}
% We evaluate the performance of different methods based on two metrics: (1) Harmonic Mean (HM) of accuracy on seen and unseen classes; (2) Incremental Class Detection Delay (ICDD).

% We evaluate the performance of different methods based on the following metrics:
% \textbf{(i) Harmonic Mean (HM) of accuracy on seen and unseen classes:} We first calculate the accuracy on seen classes and unseen classes. To equally weigh the performance of both seen and unseen classes, we calculate the harmonic mean of these two accuracy values; 
% \textbf{(ii) Incremental Class Detection Delay (ICDD):} We report the proposed ICDD metric (in \% area) specifically designed for open-world scenarios. ICDD measures the delay in discovering new classes, as described in Section~\ref{sec:metric-icdd}.
% *******************************
\subsection{Experimental Results and Analysis}

\noindent\textbf{Comparison with baseline TTA and AL methods.} In Table~\ref{tab:main-results-hm} and \ref{tab:main-results-icdd}, we observe that the performance of SegAssist is better than prior baselines in most cases (across TTA and AL baselines), only at times second to MSP. Although SegAssist tries to judiciously use the budget for unseen classes coming later in time, unseen classes semantically related to seen classes can have background ratio $B(x_t)<\alpha$. Here, MSP would select the test sample for active labeling while SegAssist would not, leading to delayed class detection in some cases. However, SegAssist performs better in majority of the cases. We report the seen and unseen class accuracy in Table~\ref{app-tab:detailed-metrics-IN-R-A} and ~\ref{app-tab:detailed-metrics-clipart-painting}. 
%Although SegAssist increases the odds for unseen sample selection, it cannot deterministically ensure improved performance, which is a {\em limitation} of this method.

% \noindent\textbf{SegAssist as a plugin module. }
% SegAssist can not only be used with MSP, but also with other standard AL techniques to further filter samples for active querying. 
% In Table~\ref{tab:segassist-acl}, we evaluate its effectiveness when used with different active learning strategies including MSP, Entropy, and Margin. We observe that SegAssist consistently improves the performance of all active learning strategies.
% \tablePluginModule

% \vspace{-10pt}
\noindent\textbf{Varying Budget rate. } We vary the budget rate as 0.5\%, 1.0\% and 1.5\% to study the effect of the budget rate (Table~\ref{sub-tab:varying-budget}) on the performance of our method. We observe that the performance of our method consistently improves with respect to the baseline AL method MSP across all budget rates. A higher budget rate naturally allows the model to query more samples, thereby enabling it to discover new classes faster, improving the overall performance.
% \tableVaryingBudget

\noindent\textbf{Varying uncertainty threshold. } We vary the threshold for MSP and SegAssist as 0.1, 0.2 and 0.3. SegAssist consistently outperforms MSP across all thresholds (Table~\ref{sub-tab:varying-threshold}).
% From Table~\ref{sub-tab:varying-threshold}, we observe that, across all thresholds chosen, SegAssist consistently outperforms MSP.
% as it judiciously selects active samples which predominantly have background pixels on performing pixel-wise segmentation. 
% This is because SegAssist uses pixel-wise segmentation to identify samples which predominantly have background pixels, relatively increasing the probability of selecting unseen class sample for querying.
% \tableVaryingThreshold

\noindent\textbf{Varying ratio of unknown to known classes.}  
We evaluate our module across test scenarios with varying unknown-to-known class ratios: 0.1, 0.25, 0.5, 0.75, and 1.0 on DN-Painting. From Table~\ref{sub-tab:varying-ratio}, we observe SegAssist consistently outperforms MSP across all ratios and methods.

% \noindent\textbf{Varying ratio of unknown to known classes. }
% To evaluate the effectiveness of our module across test scenarios where the ratio of unknown to known classes vary, we experiment with ratios 0.1, 0.25, 0.5, 0.75 and 1.0 on DN-Painting dataset, keeping the known classes fixed to 100. From Table~\ref{sub-tab:varying-ratio}, we observe that SegAssist consistently outperforms MSP across all ratios and methods.
% thereby demonstrating its effectiveness in handling varying ratios of unknown to known classes.
% \tableVaryingRatio

%\ToDo{Why top5? change and see?}

% \vspace{-5pt}
\noindent\textbf{Scope of this work.} Our goal is to formalize the ITTA problem and establish strong baselines using existing Active Labeling and TTA methods, followed by proposing a simple solution for ITTA. We introduce a lightweight plugin module, SegAssist, that can be integrated with any existing TTA method, which is the primary focus of this work. Our goal is {\em not} to introduce a new TTA method, and our approach is {\em not} positioned as a competitor to large models like GPT-4; rather, we focus on lightweight, efficient adaptation of accessible models like CLIP for real-world deployment.  

\noindent\textbf{Future research directions.}
In this work, actively labeled samples are only used to incorporate new classes. These samples could further enhance model performance through finetuning or to guide diverse sample selection for active labeling. Developing advanced active labeling methods and specialized ITTA algorithms is an important direction for future research but beyond our scope.

% \noindent\textbf{Scope for future research. } 
% In the current framework, the actively labeled samples are only utilized for incorporating new classes. These uncertain labeled samples can be further leveraged to improve model performance, through say finetuning. The labeled samples can also be used to choose diverse samples for active labeling for improved performance.  

\section{Conclusion}
\label{sec:conclusion}

Towards the goal of enabling VLMs to continuously adapt in real-world scenarios often characterized by both covariate and label shifts, we introduce and setup a benchmark for Incremental Test-Time Adaptation (ITTA).
%We leverage VLMs with active labeling to achieve this. 
We propose {\em SegAssist}, a simple plug-in module which can be seamlessly integrated with current baseline TTA methods, to prioritize active labeling of unseen class samples over seen classes. Through extensive experiments, our results demonstrate that the proposed ITTA framework, enhanced by SegAssist, advances the adaptability of VLMs in dynamic, real-world settings. We hope this work will pave the way for further research in adapting VLM-based systems in an open-world.

{
    \small
    \bibliographystyle{ieeenat_fullname}
    \bibliography{main}
}

\appendix
\section{Appendix}
\subsection{SegAssist as a plugin module. } 

SegAssist can not only be used with MSP, but also with other standard AL techniques to further filter samples for active querying. 
We evaluate its effectiveness when using Entropy measure to initially select uncertain samples (Table~\ref{tab:varying-entropy-threshold}). 
% \noindent\textbf{Varying entropy threshold.}
Further, we vary the threshold for Entropy based selection and SegAssist as 0.4, 0.5 and 0.6. From Table~\ref{tab:varying-entropy-threshold}, we observe that, for threshold 0.4 and 0.5, using SegAssist in addition to Entropy outperforms simple Entropy based active selection. For 0.6 threshold, we observe that test samples were not being selected, suggesting that it is a very high entropy threshold and hence most samples fail to qualify this criterion, making the active labeling strategy practically inactive here. Hence, no new classes are recognized resulting in a HM of almost 0 in this case. This also suggests that MSP is a better measure for active sample selection as it is more robust across the choice of MSP thresholds as shown in Table 4b. in the main paper.
% This is because SegAssist uses pixel-wise segmentation to identify samples which predominantly have background pixels, relatively increasing the probability of selecting unseen class sample for querying.
\begin{table}[h]
    \begin{adjustbox}{max width=\linewidth}
    \centering
\begin{tabular}{@{}cccccc@{}}
    \toprule
    \multirow{2}{*}{Threshold} & \multirow{2}{*}{Method} & \multicolumn{2}{c}{DN-Painting} & \multicolumn{2}{c}{ImageNet-R} \\
    \cmidrule(l){3-4} \cmidrule(l){5-6}
                            &                         & HM$\uparrow$ & ICDD $\downarrow$ & HM$\uparrow$ & ICDD $\downarrow$          \\
    \midrule
    \multirow{2}{*}{0.4}       & Entropy                  & 69.80 & 40.35 & 60.22 & 18.23      \\
                               & +SegAssist               & 69.88 & 39.83 & \textbf{62.07} & \textbf{16.15}      \\
    \midrule
    \multirow{2}{*}{0.5}       & Entropy                  & 62.87 & 45.22  & 59.88 & 21.99        \\
                               & +SegAssist               & 64.34 & 43.89 & \textbf{60.21} & \textbf{21.66}       \\
    \midrule
    \multirow{2}{*}{0.6}       & Entropy                  & 0.0  & 81.81 & 0.13 & 56.05      \\
                               & +SegAssist               & 0.0  & 81.81 & 0.13 & 56.05      \\
    \bottomrule
\end{tabular}
    \end{adjustbox}
\caption{Varying entropy threshold for active sample selection.}
\label{tab:varying-entropy-threshold}
\end{table}

\subsection{Varying topK classes for SegAssist}

In this analysis, we vary the number of topK classes to select the foreground classes $C_{seg}$ in order to perform pixelwise segmentation in SegAssist. From Table~\ref{tab:varying-topk-segassist}, we observe that on selection either 3, 5 or 7 most confident classes, the performance of SegAssist is very similar. This is because, the segmentation maps could be very similar irrespective of the choice of K for low confident samples, as majority of the samples tend to be background images (most pixels classified as background) as shown from the statistics presented in Fig. 2(b) in the main paper. However, SegAssist consistently outperforms MSP for all choices of topK classes, suggesting that the performance of SegAssist is not sensitive to the choice of K.

\begin{table}[h]
    \begin{adjustbox}{max width=\linewidth}
    \centering
\begin{tabular}{@{}cccccc@{}}
    \toprule
    \multirow{2}{*}{topK} & \multirow{2}{*}{Method} & \multicolumn{2}{c}{DN-Painting} & \multicolumn{2}{c}{ImageNet-R} \\
    \cmidrule(l){3-4} \cmidrule(l){5-6}
                            &                         & HM$\uparrow$ & ICDD $\downarrow$ & HM$\uparrow$ & ICDD $\downarrow$          \\
    \midrule
        N/A & MSP     & 70.59 & 39.14 & 62.92 & 17.21 \\ \midrule
        3 & SegAssist & 70.73 & 41.25 & \textbf{64.30} & \textbf{15.95} \\
        5 & SegAssist & 71.21 & 38.56 & 64.05 & 16.01 \\ 
        7 & SegAssist & \textbf{72.11} & \textbf{37.65} & 64.00 & 16.10 \\ 
    \bottomrule
\end{tabular}
    \end{adjustbox}
\caption{Varying top$K$ classes for selecting $C_{seg}$ in SegAssist.}
\label{tab:varying-topk-segassist}
\end{table}

\begin{table*}[t]
\begin{adjustbox}{max width=\linewidth}
\footnotesize
% \begin{tabular}{lccccccccccccc}
\begin{tabular*}{\linewidth}{@{\extracolsep{\fill}}cccccccccccccc@{}}
\toprule
& \multirow{2}{*}{Method} & \multicolumn{4}{c}{ImageNet-R} & \multicolumn{4}{c}{ImageNet-A} \\
 \cmidrule(r){3-6} \cmidrule(r){7-10} 
&  & $Acc_S\uparrow$ & $Acc_U\uparrow$ & HM$\uparrow$ & ICDD $\downarrow$ &  $Acc_S\uparrow$ & $Acc_U\uparrow$ & HM$\uparrow$ & ICDD $\downarrow$    \\ \midrule

\multirow{4}{*}{ZS-Eval} & Random    & 76.39          & 36.69          & 49.57          & 25.90           & 52.02          & 15.26          & 23.60           & 45.33          \\
                         & Entropy   & 77.21          & 48.91          & 59.89          & 21.99          & 52.41          & 13.22          & 21.11          & 43.83          \\
                         & MSP       & 76.98          & 53.2           & 62.92          & 17.21          & \textbf{52.24} & \textbf{16.08} & \textbf{24.59} & \textbf{41.47}          \\
                         \cmidrule{2-10}
                         & SegAssist & \textbf{76.99} & \textbf{54.83} & \textbf{64.05} & \textbf{16.01} & 52.24          & 15.53          & 23.94          & 41.89 \\
\midrule
\multirow{4}{*}{TDA}     & Random    & 80.2           & 39.95          & 53.33          & 25.90           & 55.19          & 8.51          & 16.58 & 45.33          \\
                         & Entropy       & 80.67          & 53.12          & 64.06          & 19.52          & 55.78         & 11.29 & 18.78          & 44.95          \\
                         & MSP   & 80.53          & 62.5           & 70.38          & 13.25          & 55.75          & 11.65          & 19.27          & 43.93          \\
                         \cmidrule{2-10}
                         & SegAssist & \textbf{80.53} & \textbf{62.66} & \textbf{70.48} & \textbf{13.07} & \textbf{55.84}          & \textbf{12.19} & \textbf{20.02} & \textbf{43.93}  \\
\midrule
\multirow{4}{*}{DPE}     & Random    & 79.47          & 41.94          & 54.90           & 25.90           & 54.26          & 19.35          & 28.52          & 45.33          \\
                         & Entropy       & 80.75          & 0.1            & 0.19           & 56.05          & 54.97          & 2.32           & 4.44           & 57.88          \\
                         & MSP   & \textbf{80.05} & 59.24          & 68.09          & 19.64          & \textbf{54.84} & 15.87          & 24.62          & 47.18          \\
                         \cmidrule{2-10}
                         & SegAssist & 79.99          & \textbf{61.28} & \textbf{69.39} & \textbf{16.38} & 54.43          & \textbf{24.11} & \textbf{33.42} & \textbf{44.98} \\
\bottomrule

\end{tabular*}
\end{adjustbox}
\caption{Detailed Metrics for ImageNet-R and ImageNet-A datasets.}
\label{app-tab:detailed-metrics-IN-R-A}
\end{table*}

\begin{table*}[t]
\begin{adjustbox}{max width=\linewidth}
\footnotesize
% \begin{tabular}{lccccccccccccc}
\begin{tabular*}{\linewidth}{@{\extracolsep{\fill}}cccccccccccccc@{}}
\toprule
& \multirow{2}{*}{Method} & \multicolumn{4}{c}{Clipart} & \multicolumn{4}{c}{Painting} \\
 \cmidrule(r){3-6} \cmidrule(r){7-10} 
&  & $Acc_S\uparrow$ & $Acc_U\uparrow$ & HM$\uparrow$ & ICDD $\downarrow$ &  $Acc_S\uparrow$ & $Acc_U\uparrow$ & HM$\uparrow$ & ICDD $\downarrow$    \\ \midrule
\multirow{4}{*}{ZS-Eval} & Random    & 83.21          & 12.71          & 22.04          & 75.30          & 81.71          & 20.31          & 32.54          & 68.41          \\
                         & Entropy   & 83.79          & 33.85          & 48.25          & 61.88          & 81.59          & 53.1           & 64.34          & 43.89          \\
                         & MSP       & 83.74          & 33.3           & 47.65          & 62.04          & 81.49          & 62.27          & 70.59          & 39.14          \\
                         \cmidrule{2-10}
                         & SegAssist & \textbf{83.67} & \textbf{34.72} & \textbf{49.08} & \textbf{61.2}  & \textbf{81.59} & \textbf{63.17} & \textbf{71.21} & \textbf{38.56} \\
\midrule
\multirow{4}{*}{TDA}     & Random    & 83.25          & 13.03          & 22.54          & 75.30           & 81.25          & 20.43          & 32.66          & 68.41          \\
                         & Entropy   & 83.56          & 45.35          & 58.79          & 49.1           & 81.28          & 57.38          & 67.27          & 46.08          \\
                         & MSP       & 83.63          & 45.56          & 58.99          & 50.68          & 81.29          & 59.67          & 68.83          & \textbf{39.31} \\
                         \cmidrule{2-10}
                         & SegAssist & \textbf{83.61} & \textbf{47.10} & \textbf{60.25} & \textbf{48.96} & \textbf{81.32} & \textbf{60.22} & \textbf{69.2}  & 41.27          \\
\midrule
\multirow{4}{*}{DPE}     & Random    & 83.45          & 13.58          & 23.36          & 75.30          & 81.19          & 21.10          & 33.49          & 68.41          \\
                         & Entropy   & 83.67          & 8.54           & 15.5           & 80.27          & 81.44          & 0              & 0              & 81.81          \\
                         & MSP       & 83.52          & 53.67          & 65.35          & 49.00          & 80.07          & 64.5           & 71.45          & 34.7           \\
                         \cmidrule{2-10}
                         & SegAssist & \textbf{83.47} & \textbf{53.89} & \textbf{65.49} & \textbf{49.00} & \textbf{80.00} & \textbf{66.12} & \textbf{72.40} & \textbf{33.12} \\
\bottomrule
\end{tabular*}
\end{adjustbox}
\caption{Detailed Metrics for Clipart and Painting domains of DomainNet dataset.}
\label{app-tab:detailed-metrics-clipart-painting}
\end{table*}

\subsection{\bf Empirical Analysis of ICDD and HM.}
\label{app:metrics-discussion}
We report the detailed metrics of the experimental results (Table 2. and 3. in the main paper) including Accuracy on seen ($Acc_S$) and unseen ($Acc_U$) classes along with their Harmonic Mean (HM) and ICDD in Table ~\ref{app-tab:detailed-metrics-IN-R-A} and~\ref{app-tab:detailed-metrics-clipart-painting}.
To further illustrate the complementary nature of ICDD and HM, we analyze a few specific cases which demonstrate that similar ICDD values do not necessarily imply similar accuracy, and vice versa, reinforcing the need for both metrics in evaluating Incremental Test-Time Adaptation (ITTA).

\begin{itemize}
    \item {\bf Same ICDD, Different HM}:  
    For the Random strategy, since the test stream remains the same and the same random seed is used, the selected test samples for active labeling remain consistent across different TTA methods. As a result, the ICDD values remain the same for all TTA methods under the Random selection strategy. However, the classification performance varies significantly across methods. Notably, TDA and DPE, being more effective TTA methods, achieve higher HM values compared to ZS-Eval, demonstrating that even when detection delay remains unchanged, the choice of TTA method plays a crucial role in improving classification accuracy.
    
    \item {\bf Similar ICDD, Different HM }:  
    On the ImageNet-R dataset, ZS-Eval and DPE (using SegAssist) exhibit similar ICDD values (16.01 and 16.38, respectively). However, their HM values differ significantly (64.05 for ZS-Eval vs. 69.39 for DPE). This highlights that while both methods detect new classes at a comparable pace, DPE yields superior classification accuracy. This scenario confirms that ICDD alone does not capture classification correctness, which is reflected by HM.

    \item {\bf Different ICDD, Similar HM}:  
    For the Painting domain, we observe that ZS-Eval and DPE achieve comparable HM values (71.21 and 72.40, respectively), indicating similar classification effectiveness for detected classes. However, their ICDD values differ significantly (38.56 for ZS-Eval vs. 33.12 for DPE), suggesting that while both methods classify detected classes with similar accuracy, their timeliness in detecting new classes varies. Specifically, DPE exhibits a lower ICDD, implying that it has detected new classes earlier compared to ZS-Eval. We attribute the reason for this case of Similar HM despite of different ICDD as follows: (1) Reduced $Acc_S$ of DPE compared to ZS-Eval. (2) Accuracy also depends on both TTA Method and the particular classes detected. More difficult classes detected in one method can result in lower accuracy compared to another method where maybe easier classes are detected. This further highlights how ICDD captures detection timeliness independently of classification performance.
\end{itemize}

\noindent These observations confirm that ICDD and HM capture different aspects of the detection process. While ICDD reflects detection timeliness, HM quantifies classification reliability. Together, they provide a comprehensive evaluation of ITTA performance, ensuring both aspects are considered when designing and comparing adaptation strategies.

\end{document}